\documentclass[runningheads]{llncs}

\usepackage[mobile]{eccv}

\usepackage{eccvabbrv}

\usepackage{graphicx}
\usepackage{booktabs}
\usepackage{wrapfig}
\usepackage{colortbl}
\usepackage{tabularx}
\usepackage[accsupp]{axessibility}  % Improves PDF readability for those with disabilities.

\usepackage{hyperref}

\usepackage{orcidlink}

\begin{document}

% \newgeometry{bottom=1cm}
\title{SGD: Street View Synthesis with Gaussian Splatting and Diffusion Prior} 

% \titlerunning{SGD}

% Second Author\inst{2,3}\orcidlink{1111-2222-3333-4444} \and
% Third Author\inst{3}\orcidlink{2222--3333-4444-5555}}
\author{Zhongrui Yu\inst{1}$^\dagger$ \and Haoran Wang\inst{2}$^\ddagger$ \and Jinze Yang\inst{3} \and Hanzhang Wang\inst{4} \and Zeke Xie\inst{2} \and Yunfeng  Cai\inst{2} \and Jiale Cao \inst{5} \and Zhong Ji\inst{5} \and Mingming Sun\inst{2}}

\authorrunning{Z.~Yu et al.}

% \institute{ETH Zurich, Switzerland \and
% Springer Heidelberg, Tiergartenstr.~17, 69121 Heidelberg, Germany
\institute{
$^1\,$ETH Zürich,
$^2\,$Baidu Research,
$^3\,$University of Chinese Academy of Sciences,
$^4\,$Harbin Institute of Technology,
$^5\,$Tianjin University\\
\email{zhonyu@ethz.ch, 
wanghaoran09@baidu.com
% \{wanghaoran09, xiezeke, caiyunfeng, sunmingming01\}@baidu.com
% yangjinze20@mails.ucas.ac.cn, cswhz@hit.edu.cn, \\
% \{connor, jizhong\}@tju.edu.cn
}}

\maketitle

\vspace{-0.9cm}

%TODO: caption
\begin{figure}[h]
  \centering
  \includegraphics[width=0.95\linewidth]{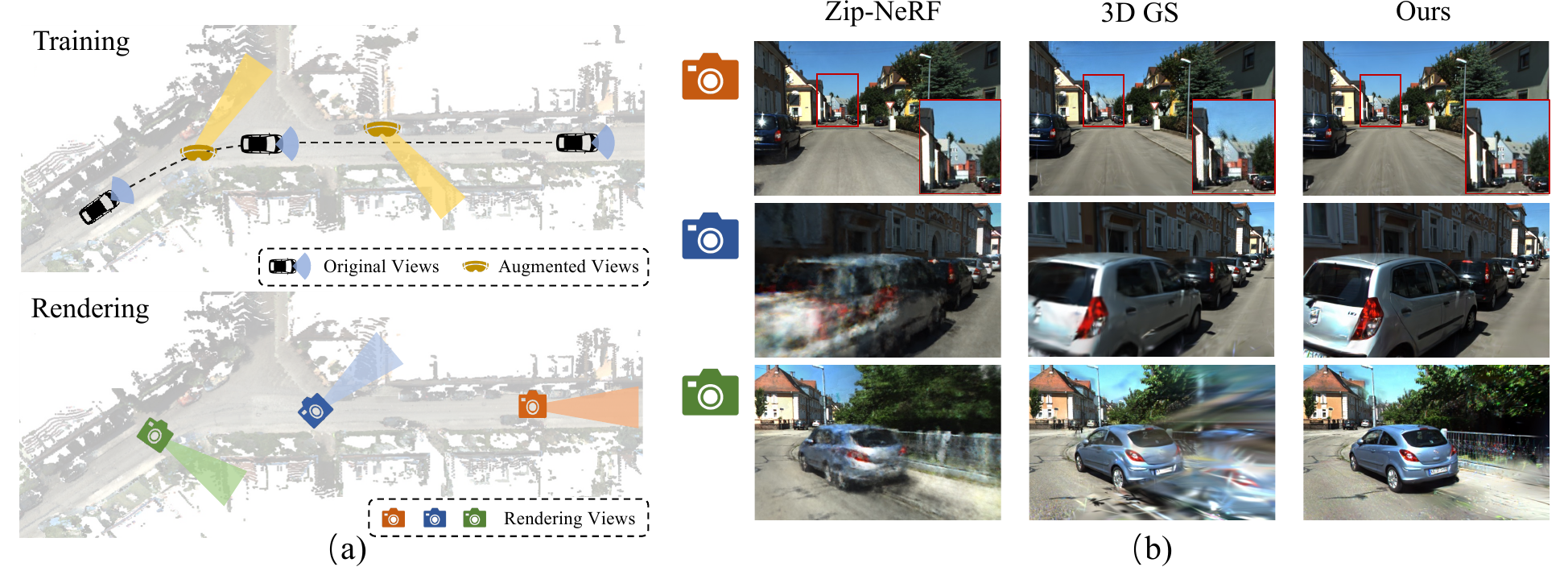}
  \caption{\textbf{(a)}. 
  % The goal of employing Novel View Synthesis (NVS) methods in autonomous driving simulation is to render images from any desired viewpoints, enabling the free control of ego-vehicle. 
  To enable free control of ego-vehicle in autonomous driving simulation with novel view synthesis, we propose a method that leverages the prior from a Diffusion Model to provide 3DGS\cite{3dgs} augmented views during training. \textbf{(b)}. Our method preserves photo-realistic rendering quality at viewpoints that are distant from the training views while other approaches\cite{barron2023zipnerf, 3dgs} produce severe artifacts.
  }
  \label{fig:1}
\end{figure}

\vspace{-1.4cm}

\renewcommand{\thefootnote}{\fnsymbol{footnote}}
\footnotetext[4]{Work done when Zhongrui Yu was a Research Intern at Baidu Research.}
\footnotetext[5]{The corresponding author.}

%%%%%%%%%%%%%%%%%%%%%%%%%%abstract
\begin{abstract}

%%%%%v3
%%%%%v3
Novel View Synthesis (NVS) 
% methods, including Neural Radiance Fields (NeRF) and 3D Gaussian Splatting (3DGS), 
for street scenes
play a critical role in the autonomous driving simulation.
The current mainstream technique to achieve it is neural rendering, such as Neural Radiance Fields (NeRF) and 3D Gaussian Splatting (3DGS). 
Although thrilling progress has been made, when handling street scenes, current methods struggle to maintain rendering quality at the viewpoint that deviates significantly from the training viewpoints. This issue stems from the sparse training views captured by a fixed camera on a moving vehicle. To tackle this problem, we propose a novel approach that enhances the capacity of 3DGS
% by offering augmented views, which is achieved 
by leveraging prior from a Diffusion Model along with complementary multi-modal data. Specifically, we first fine-tune a Diffusion Model by adding images from adjacent frames as condition, meanwhile exploiting depth data from LiDAR point clouds to supply additional spatial information. Then we apply the Diffusion Model to regularize the 3DGS at unseen views during training. Experimental results validate the effectiveness of our method compared with current state-of-the-art models, and demonstrate its advance in rendering images from broader views. 

% \keywords{autonomous driving \and novel view synthesis \and 3D reconstruction}
\end{abstract}

%%%%%%%%%%%%%%%%%%%%%%%%%% Introduction
\section{Introduction}
\label{sec:intro}

%%%%% 1 - The importance of autonomous driving simulation
%%% Autonomous driving simulations hold crucial importance for the technological development of autonomous vehicles. By simulating driving scenes, it becomes possible to acquire data that are otherwise difficult to gather from real driving scenarios, such as safety-critical scenarios or corner cases, without the risk of real-world damage. Simulated data contributes to tasks including perception, planning and decision-making, thereby significantly improving the safety and reliability of autonomous driving systems.

The driving simulation in street scenes holds crucial importance in the development of autonomous driving systems.
% , which benefits numerous relevant applications, such as machine perception, planning and decision-making. 
Through constructing a digital twin of urban streets, we can continually enhance our autonomous driving system with simulated data. Thereby, the dependence of data collection in real scenarios is significantly reduced, making it possible to build a powerful autonomous driving system with lower time and financial cost.

%%%%% 2 - Autonomous driving simulation with novel view synthesis methods.
% - why is nvs for street scene a sparse-view-reconstruction problem
%%% Traditional driving simulation engines\cite{carla, airsim...} often struggle with the time-consuming virtual scene construction, alongside issues of low visual realism and lack of texture detail. Recent studies\cite{} have applied Novel View Synthesis (NVS) methods, such as Neural Radiance Fields (NeRF) \cite{nef} and 3D Gaussian Splatting (3DGS)\cite{3dgs}, on autonomous driving simulation and achieved thrilling progress. The typical approach is to train either a NeRF or 3DGS model using images collected by a real-world vehicle, then utilize it within the simulation system to render the street views from desired viewpoints, enabling functionalities such as rotation and lane alteration of ego-vehicle. 

For autonomous driving simulation, the early attempts \cite{carla, shah2018airsim, rong2020lgsvl} deploy Computer Graphics (CG) engines to render the images. It not only requires the time-consuming process to reconstruct virtual scenes, but also yields results with low realism and fidelity. 
% Recently, the prevalence of neural rendering techniques, such as Neural Radiance Fields (NeRF) \cite{mildenhall2021nerf} and 3D Gaussian Splatting (3DGS)\cite{3dgs}, are introduced by more studies for street view synthesis. 
Recently, neural rendering techniques for Novel View Synthesis (NVS), such as Neural Radiance Fields (NeRF) \cite{mildenhall2021nerf} and 3D Gaussian Splatting (3DGS)\cite{3dgs}, are introduced for synthesizing photo-realistic street views. Current studies\cite{tancik2022block, rematas2022urf, lu2023dnmp, nsg, turki2023suds, wu2023mars, guo2023streetsurf, yan2024street, zhou2023drivinggaussian} mainly investigate two challenges faced in street view synthesis: the reconstruction of unbounded scenes and the modeling of dynamic objects. BlockNeRF\cite{tancik2022block} proposed to split scenes into multiple blocks, aimed at enhancing the model's capacity to present large unbounded street scenes. NSG\cite{nsg} and some following methods \cite{turki2023suds, snerf, zhou2023drivinggaussian, yan2024street, wu2023mars} separately model the static background and the dynamic foreground to achieve higher background rendering quality while reducing motion blur associated with foreground vehicles.

Although thrilling progress has been made, a critical problem for evaluating the reconstruction quality is not well explored in existing works. It is known that an ideal scene simulation system should have the capacity to achieve free-view rendering with high quality. The current works commonly adopt the views that are sourced from vehicle captures yet unseen in the training stage as {\bf test views}, (such as the red viewpoint in \cref{fig:1}), while neglecting the {\bf novel views} that deviate from the training views (such as the blue and green viewpoints in \cref{fig:1}). When handling these novel views, there is a noticeable reduction in rendering quality with blurring and artifacts for existing works, as shown in \cref{fig:1}. This issue is attributed to the inherently constrained view of the vehicle-collected images. The training images are typically captured along the vehicle's traveling direction and centered around the vehicle's lane. Due to the fast traveling speed of the vehicle, there is limited overlap between frames, thus not allowing for comprehensive multi-view observation of objects in the scene. Therefore, the street view synthesis task for autonomous driving can be comprehended as a reconstruction problem from sparse views.

% 3 - Meanwhile, NVS methods with sparse inputs are developed: mostly for object-centric scenes & single-object generation; 
% - Why Diffusion models? Why not pose-conditioned Diffusion Model?
% - 
% In the meantime, several methods have been proposed to address the challenge of NVS from sparse views.  The emergence of single image-to-3D generation methods\cite{} offers new insight to leverage a pre-trained Diffusion Model for NVS. They typically fine-tuned a text-to-image Diffusion Model on some large multi-view datasets \cite{shapenet, objaverse, mvimgnet, co3d} to an image-to-image Diffusion Model with relative camera poses as condition, subsequently apply the Diffusion Model to regularize the training of a NeRF model or a 3DGS model. However, in more complex scenes like street scenes, relying solely on relative camera poses is insufficient for learning the geometric details, thereby limiting the applicability of these methods to the reconstruction of a single object\cite{3dim, liu2023zero} or object-centric scenes\cite{zeronvs, GeNVS}. Most recent work ReconFusion\cite{wu2023reconfusion} extends this approach to more general 3D scene reconstruction, yet it still requires a jointly trained PixelNeRF to encode relative poses. To better utilize the Diffusion Model for NVS in autonomous driving context, we seek to derive the scene's 3D geometric information from other modality data to control the Diffusion model, thus freeing it from the dependency on relative camera poses.

Previous neural rendering methods proposed to address the challenge of NVS from sparse views can be categorized into two main branches. The first branch\cite{verbin2022refnerf,deng2022dsnerf,yu2021pixelnerf,wynn2023diffusionerf,somraj2023VipNeRF} incorporates scene priors, such as depth\cite{deng2022dsnerf,roessle2022ddpnerf}, normal\cite{verbin2022refnerf}, or the features extracted from a deep network\cite{yu2021pixelnerf} to regularize the model training in an explicit manner. Besides, another branch\cite{poole2022dreamfusion,liu2023zero,shi2023zero123++,wu2023reconfusion,sargent2023zeronvs} attempts to leverage a pre-trained Diffusion Model for NVS. They typically fine-tuned a text-to-image Diffusion Model on some large multi-view datasets \cite{chang2015shapenet,deitke2023objaverse,yu2023mvimgnet,reizenstein2021co3d} to an image-to-image Diffusion Model with relative camera poses as condition, subsequently apply the Diffusion Model to regularize the training of neural rendering model. However, a significant domain gap persists between multi-view datasets\cite{chang2015shapenet,deitke2023objaverse,yu2023mvimgnet,reizenstein2021co3d} and street scenes. And relying solely on relative camera poses is insufficient to learn the geometric details in more complex street scenes. To resolve this issue in the autonomous driving context, we leverage 3D geometric information obtained from multi-modal data to control the Diffusion model, enabling direct fine-tuning it on autonomous driving datasets and getting rid of the necessity of encoding relative camera poses.

% 4 - Brief introduction of our method
To consolidate the idea, in this paper, we propose a novel NVS approach for street scenes, based on 3D Gaussian Splatting and prior from a fine-tuned Diffusion Model. We start by fine-tuning a Diffusion Model on an autonomous driving dataset\cite{liao2022kitti}. For each input image, we employ its adjacent frames as the condition and leverage the depth information from LiDAR point clouds as control. This fine-tuned Diffusion Model then aids in guiding the 3DGS training by providing prior for the unseen views. Our method demonstrates competitive performance with the state-of-the-art (SOTA) methods\cite{wu2023mars, barron2023zipnerf,3dgs} on KITTI\cite{geiger2015kitti} and KITTI-360\cite{liao2022kitti} datasets with dense viewpoint inputs and outperforms them in the sparse-view setting. Remarkably, our approach maintains high rendering qualities even for the viewpoints distant from training views. Moreover, since our approach is only applied during training, it does not compromise the real-time inference capability of 3DGS. Therefore, our model facilitates efficient rendering and versatile viewpoint control within autonomous driving simulation systems.

% 5 - Contributions
In Summary, we provide the following contributions:
\begin{itemize}
% \item We propose a novel framework for Novel View Synthesis in street scenes, enhancing the freedom of view control and sustaining rendering efficiency for autonomous driving simulations.
\item We propose a novel framework for Novel View Synthesis in street scenes, enhancing the freedom of view control in the premise of sustaining rendering efficiency for autonomous driving simulations.
% \item To the best of our knowledge, we first identify the street view synthesis problem as a sparse-view input problem, and address this challenge by integrating 3D Gaussian Splatting and Diffusion Model. 
\item To the best of our knowledge, our method is the first attempt to tackle the street view synthesis task from the perspective of sparse-view-input reconstruction problem, and address this challenge by combining 3D Gaussian Splatting with a customized Diffusion Model. 
% \item We first investigated the fine-tuning strategy of the Diffusion Model on multi-modal data from autonomous driving datasets. which overcomes the conventional reliance on multi-view data and relative camera poses, and proves its effectiveness.
\item A novel strategy for fine-tuning a Diffusion Model on autonomous driving datasets and equipping it with NVS capability is presented, which overcomes the conventional reliance on multi-view datasets and relative camera poses.
% tailored diffusion model 
% is presented for street view generation, which is capable of fully making use of multi-modal data from autonomous driving datasets. The experimental results validate its effectiveness in ameliorating the reliance on multi-view data and relative camera poses.

\end{itemize}

%%%%%%%%%%%%%%%%%%%%%%%%%% Related Work

\section{Related Work}
\label{sec:relat}

% \subsection{Novel View Synthesis Method}
% - NeRF and beyond
% - 3DGS

\subsubsection{Novel View Synthesis for Street Scenes}

% unbounded scene modeling, dynamic vehicle modeling
The rapid development of the NVS techniques including NeRF\cite{mildenhall2021nerf} and 3DGS\cite{3dgs} has attracted considerable attention within the arena of autonomous driving. A multitude of studies \cite{tancik2022block, rematas2022urf, lu2023dnmp, nsg, turki2023suds, wu2023mars, guo2023streetsurf, yan2024street, zhou2023drivinggaussian, liu2023real, chen2023periodic, tonderski2023neurad, yang2023unisim} have explored the utilization of these methods for street-view synthesis. Block-NeRF\cite{tancik2022block} and Mega-NeRF\cite{turki2022mega} have proposed to segment the scenes into distinct blocks for individual modeling. Urban Randiance Field\cite{rematas2022urf} enhances the NeRF training with geometric information for LiDAR. DNMP\cite{lu2023dnmp} utilizes a pre-trained deformable mesh primitive to represent the scene. Streetsurf\cite{guo2023streetsurf} delimits the scene into close-range, distant-view and sky, achieving superior reconstruction results for urban street surfaces. For the modeling of dynamic urban scenes, NSG\cite{nsg} presents the scene as neural graphs. MARS\cite{wu2023mars} employs distinct networks for modeling background and vehicles, creating an instance-aware simulation framework. With the emergence of 3DGS\cite{3dgs}, DrivingGaussian\cite{zhou2023drivinggaussian} introduces Composite Dynamic Gaussian Graphs and incremental static Gaussians. StreetGaussian\cite{yan2024street} optimizes the tracked pose of dynamic Gaussians and introduces 4D SH (spherical harmonics) for varying vehicle appearance in different frames.

In summary, current methods for street view synthesis mainly focus on two challenges: reconstructing the large-scale unbounded scenes and accurately modeling the dynamic vehicles. Yet the sparse-view-input issue within this task has not been adequately addressed.

\subsubsection{Novel View Synthesis with Sparse View Inputs}

For NVS methods\cite{mildenhall2021nerf, 3dgs}, intensive capture of the scene is paramount. Ideally, each part of the scene should be observed from serval perspectives, and a large overlap should exist across frames. The challenge of capturing such data leads to the development of methods that aim at improving rendering quality with sparse inputs\cite{verbin2022refnerf, deng2022dsnerf,yu2021pixelnerf,sargent2023zeronvs, wu2023reconfusion, yang2023freenerf, wynn2023diffusionerf, roessle2023ganerf, wang2023sparsenerf, zhou2023sparsefusion, chan2023genvs, xiong2023sparsegs, kwak2023geconerf, somraj2023VipNeRF}. Early approaches involve supplementing the training process with scene priors. RefNeRF\cite{verbin2022refnerf} incorporates a pre-trained model for normal flow to regularize novel viewpoints. DS-NeRF\cite{deng2022dsnerf} enhances the reconstruction with depth information derived from SfM (Structure from Motion) point clouds. PixelNeRF\cite{yu2021pixelnerf} leverages a CNN encoder to extract features from images. The encoder can be trained across different scenes to acquire diverse priors. 

Serval current methods leverage prior from Diffusion Model, which is pre-trained on large-scale datasets, to support the synthesis of novel views. Zero-1-to-3\cite{liu2023zero} and ZeroNVS\cite{sargent2023zeronvs} fine-tune the Diffusion Model by conditioning it on single image and a relative camera pose. Zero123++\cite{shi2023zero123++} develops various conditioning and training schemes to minimize the effort of fine-tuning. ReconFusion\cite{wu2023reconfusion} jointly train a PixelNeRF\cite{yu2021pixelnerf} along with fine-tuning the Diffusion Model, with the feature extracted by PixelNeRF serving as conditions for the Diffusion Model. It proves that PixelNeRF's feature provides a more accurate representation of the relative camera pose. However, the fine-tuning process of these methods is typically on large multi-view datasets, including ShapeNet\cite{chang2015shapenet}, Objaverse\cite{deitke2023objaverse}, CO3D\cite{reizenstein2021co3d}, MVImgNet, \etc. These datasets are object-centric, maintaining a large domain gap from driving scenes. Inspired by the aforementioned methods, we propose a novel fine-tuning approach for Diffusion model tailored to autonomous driving scenarios, leveraging the 3D spatial information provided by multimodal data as conditions.

%%%%%%%%%%%%%%%%%%%%%%%%%% Method
\section{Method}
\label{sec:method}

The goal of street view synthesis is to render images from any viewpoint $v$, given a set of images and corresponding camera poses $\{I_i, \Vec{p}_i\}_{i=1}^N$ captured by a vehicle. A big challenge arises from the constrained perspectives of images collected by a moving vehicle, where objects in the scene are often only observed from single viewpoint and appear only in a few images. To address this issue, we propose a novel method that leverages the priors derived from a fine-tuned Diffusion Model and the spatial information from LiDAR, to enhance the 3DGS model's awareness of the unobserved world.

Our method consists of two main components. We first fine-tune a Stable Diffusion Model\cite{rombach2022sd} on a dataset of driving scenes \cite{liao2022kitti} conditioning on reference images from adjacent frames and depth from LiDAR point cloud (detailed in \cref{subsec:fine-tuneSD}). Subsequently, we integrate the fine-tuned Diffusion Model into the 3D Gaussian Splatting pipeline to guide the synthesis of unseen views (detailed in \cref{subsec:3dgs}).
% \subsection{Preliminaries}

\begin{figure}[tb]
  \centering
  \includegraphics[width=0.98\linewidth]{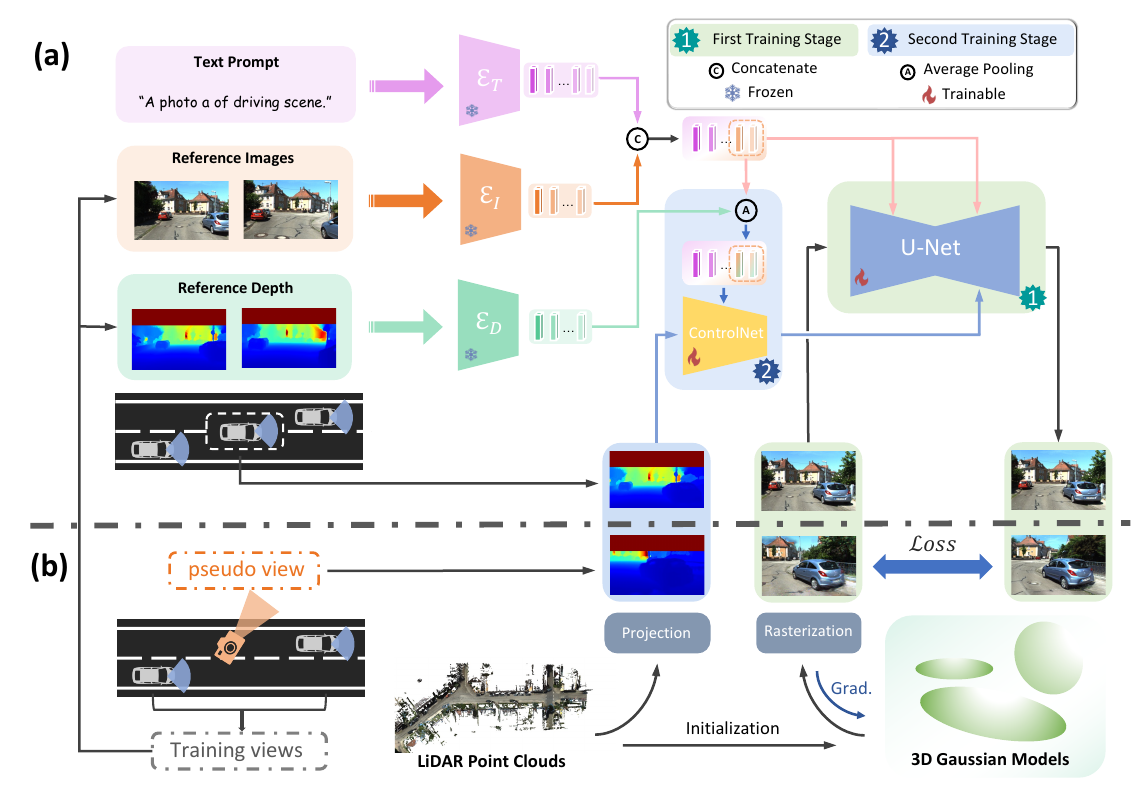}
  
  \caption{\textbf{Overview of Our Method.} \textbf{(a).} There are two training stages in the Diffusion Model\cite{rombach2022sd} fine-tuning. Firstly, the U-Net is fine-tuned by being injected with the patch-wise CLIP Image features of reference images concatenated with the CLIP text features of a text prompt. Secondly, a ControlNet is trained with the depth of the target image as the control signal. \textbf{(b).}The fine-tuned Diffusion Model from (a) guides the 3DGS training by providing regularization in pseudo views. For the sake of simplicity, the VAE encoder and decoder are omitted in the figure.
  }
  %TODO:
  \label{fig:method}
\end{figure}

\subsection{Fine-tuning Diffusion model}
\label{subsec:fine-tuneSD}
% problem formulation
% The remarkable capability of Diffusion Model in image generation comes from its ability to learn data distributions. Many previous works \cite{poole2022dreamfusion, liu2023zero} have shown that Diffusion Model is capable of learning the conditional distribution of an object in various viewpoints based on a single viewpoint image and its relative camera poses. This typically involves the explicit representation of relative position (\ie, translation and rotation) as the condition for Diffusion Model\cite{liu2023zero}. Such strategy is usually limited to small-scale scenes or object-centric scenes. Recent work ReconFusion\cite{wu2023reconfusion}, utilizes implicit features to represent relative positions, more specifically the hidden features of a jointly trained PixelNeRF\cite{yu2021pixelnerf}. However, the storage capabilities of PixelNeRF limit its applicability in complex, large-scale scenes.

We propose a novel approach to fine-tune a Diffusion Model specifically on driving data. Driving data is collected sequentially, allowing us to easily identify the closest preceding and succeeding frames from any novel viewpoint. The images from these adjacent frames are taken as reference images as they offer valuable contextual information. Moreover, the 360° LiDAR point clouds allow us to derive depth maps for both reference frames and the novel view, offering a comprehensive understanding of the relative spatial information across viewpoints. Briefly, by fine-tuning the Diffusion model, we guide it to learn about the contents that ought to be present from the context images and the spatial relationships among objects from the depth information.

%TODO: fig & refine
The structure of our model is illustrated in \cref{fig:method} (a). 
% The input image is firstly encoded by a variational auto-encoder (VAE) to a latent space. Then noise is added to the latent features by the noise scheduler. 
% The U-Net of the Diffusion Model takes the latent feature map extracted from a training image as input, and conducts a multi-step denoising process on the noisy latents. 
% And then a VAE decoder decodes the latents to recover the original image. 
During fine-tuning, information from reference images is introduced via cross-attention within the U-Net (in the first training stage), and the depth information is introduced by a ControlNet\cite{zhang2023control}-like module (in the second training stage).

\subsubsection{Training Stage 1: Image Conditioned Diffusion Model}
In the first step, we want the Diffusion Model to learn high-level information about the scene from images of adjacent frames. Unlike other methods, such as ReconFusion\cite{wu2023reconfusion}, which introduces images and poses as conditions at the same time, we opt to not include any pose information at this stage. The reason we conduct this two-step training strategy is that the driving scene is complex, containing a variety of objects including buildings, vehicles, pedestrians, \etc, and the objects have mutual occlusion. The Diffusion Model is challenged in fully understanding the 3D scene by only encoding the relative camera pose. Thus, in this step we focus on enabling the Diffusion Model to identify \emph{"what exits within the scene"}, and in next step we utilize the LiDAR point clouds to tell the model \emph{"where each object is located"}. 
% TODO: in ablation study we shown the result of joint training

As depicted in \cref{fig:method}, besides the original structure of the Stable Diffusion Model\cite{rombach2022sd}, we introduce an additional pathway into the U-Net to integrate information from reference images encoded by the CLIP Image Encoder\cite{radford2021clip}. During the fine-tuning, We freeze the VAE encoder, decoder, CLIP Text Encoder and CLIP Image Encoder, and keep the parameters of U-Net $\theta$ trainable. The input image $I$ is encoded as a latent feature map $z_0$ by the VAE encoder. A text prompt $T$ is tokenized and encoded by the CLIP Text Encoder $\mathcal{E}_{text}$ to obtain text embedding,
\begin{equation}
  e_T = \mathcal{E}_{\text{text}}(T) = [e_T^0, \ldots, e_T^{L_T}], \quad e_T^i \in \mathbb{R}^D
  \label{eq:textencoder}
\end{equation}
where $L_T$ denotes the length of the text token sequence and $D$ denotes the dimension of the CLIP embeddings. For each input image $I \in \mathbb{R}^{3\times H \times W}$, a previous and a next frame are selected as the reference images $I_\text{ref} = \{I_{\text{pre}}, I_\text{next}\}$. Similar to the pathway of text, the reference images $I_\text{pre}$ and $I_\text{next}$ are encoded by the CLIP Image Encoder\cite{radford2021clip} $\mathcal{E}_\text{Image}$ separately, and the latent patch-wise CLIP Image features $e_{I_\text{ref}}=[e_{I_\text{pre}}, e_{I_\text{next}}]$ are concatenated to the text embedding along the dimension of token length. 
\begin{equation}
\label{eq:imageencoder}
  e_{I_i} = \mathcal{E}_{\text{Image}}(I_i) = [e_{I_i}^0, \ldots, e_{I_i}^{L_I}], \quad i \in \{\text{pre}, \text{next}\}
\end{equation}
where $L_I$ indicates the number of image patches.
Through this process, we conduct a sequence of embeddings including not only the coarse-level information of the scene conveyed by the text prompt, but also the detailed semantic information introduced by the adjacent frames. This embedding is fused with the feature map $z_0$ via the cross-attention layer in the U-Net. The fine-tuning process is optimized by the loss funcion:
\begin{equation}
     L({\theta}) = \mathbb{E}_{z_0, \epsilon \sim \mathcal{N}(\mathbf{0}, \mathbf{1}), t, T, I_\text{ref}}||\epsilon - \epsilon_\theta(z_t, t, e_T, e_{I_\text{ref}})||_2^2
\end{equation}
where $\epsilon$ is the random noise, $t$ is the denoising timestep and $z_t$ is the noisy latent at timestep $t$.   
The efficacy of the first step of fine-tuning is shown in \cref{fig:ablation-1} in the ablation study. %TODO, Fig in ablation study

\subsubsection{Training Stage 2: Adding Depth ControlNet}
The model acquires a high-level understanding of the scene after the first training step, yet it remains unaware of the spatial relationships between objects within the scene. In the second training step, we intend to utilize 3D information to control the model to achieve more accurate image generation. In autonomous driving scenarios, the 3D structure of the scene is preserved through LiDAR point clouds. Given an arbitrary viewpoint, the depth map can be derived by projecting the point clouds onto the image plane and then completed by an off-the-shelf depth completion model\cite{zhang2023completionformer}.

To enable depth control, we keep the fine-tuned Diffusion Model frozen and add a ControlNet\cite{zhang2023control} module to it, as depicted in \cref{fig:method}. The ControlNet is initialized as a trainable copy the U-Net's encoder block. From the projection of the LiDAR point cloud, both the depth map for the reference images $D_{I_\text{ref}}$ and for the input image $D_I$ can be obtained. $D_{I}$ serves as the input of the ControlNet.
% Just like the original ControlNet pipeline, the depth map $D_I$ is mapped to a conditioning feature $c_D$ and added to the noisy latent $z_0$. 
To encode the depth map of the reference images $D_{I_\text{ref}}$, a depth encoder $\mathcal{E}_{depth}$ is introduced. The depth encoder is pre-trained to align the CLIP Image Encoder with contrastive learning proposed by \cite{huang2023clip2point}.
%(detailed in CLIP2Point and supplemental materials)
The encoded depth feature of the reference images is fused to their corresponding CLIP Image features via average pooling.
\begin{align}
    e_{D_i} & = \mathcal{E}_\text{Depth}(I_i)\\
    \Tilde{e}_{I_i} & = \text{AvgPool}(e_{I_i}, e_{D_i}), \quad i \in \{\text{pre}, \text{next}\}
\end{align}
The fused feature $\Tilde{e}_{I_i}$ concatenated with $e_T$ is injected into the ControlNet via cross-attention. The ControlNet is trained with the loss function:
\begin{equation}
     L(\Tilde{\theta}) = \mathbb{E}_{z_0, \epsilon \sim \mathcal{N}(\mathbf{0}, \mathbf{1}), t, T, \{I_\text{ref}, D_\text{ref}\}}||\epsilon - \epsilon_\theta(z_t, t, c_D, e_T, \Tilde{e}_{I_\text{ref}})||_2^2
\end{equation}
where $\Tilde{\theta}$ denotes the parameters of the ControlNet. The efficacy of our ControlNet is shown in \cref{fig:ablation-1} in the ablation study. %TODO, Fig in ablation study

\subsection{3D Gaussian Splatting with Diffusion Prior}
\label{subsec:3dgs}

Once the Diffusion Model is fine-tuned, we use its prior on images from unobserved views to regularize 3D Gaussian Model training.

3D GS\cite{3dgs} represents the scene as a large number of 3D Gaussian Models, each Gaussian Model is parameterized as its mean $\Vec{\mu}$, covariance $\Vec{\mathbf{\Sigma}}$, opacity $\alpha$, and spherical harmonics parameters for view-dependent RGB color $\Vec{c}$. For simplicity, we denote $\Vec{\phi}$ as the set of all trainable parameters of a single Gaussian Model. In each training step, a training view $v_o$ is sampled and rendered via differentiable rasterization. 
% \begin{equation}
%     C = \sum_{i\in N_o} \Vec{c}_i \alpha^{\prime}_i \prod_{j=1}^{i-1} (1- \alpha^{\prime}_j) 
% \end{equation}
% \begin{equation}
%     \alpha^{\prime}_j = \alpha_i \times \exp(-\frac{1}{2}(\Vec{x}^{\prime}-\Vec{\mu_j}^{\prime}) \Vec{\mathbf{\Sigma}}_j^{\prime-1} (\Vec{x}^{\prime}-\Vec{\mu}_j^{\prime}))
% \end{equation}
% where $C$ is the rendered RGB, $N_o$ is the set of the Gaussian Models which can be seen from this training view, and $\alpha^{\prime}$ is the multiplication of opacity $\alpha$ and the projected 2D Gaussian.
The parameters of the Gaussian Models corresponding to view $v_o$, denoted as $\Phi_o = \{\Vec{\phi}_i\}_{i \in N_o}$,  are optimized by the loss function in \cref{eq:reconloss}, which is a combination of RGB loss, SSIM loss and depth loss. The depth loss is computed as the $L_1$ loss between the rendered depth map $\Tilde{D}_o$ and the completed dense depth map $D_o$ from the projection of LiDAR point clouds.
\begin{equation}
\label{eq:reconloss}
    L_\text{recon}(\Phi_o) = \mathbb{E}_{v_o}[||I_o, \Tilde{I}_o||_1 + \lambda_\text{SSIM}L_\text{SSIM}(I_o, \Tilde{I}_o) + \lambda_\text{depth}||D_o, \Tilde{D}_o||_1]
\end{equation}
% \begin{equation}
%     L_\text{depth}(\Phi_o) = \mathbb{E}_{v_o}||D_o, \Tilde{D}_o)||_1
% \end{equation}

Besides the training view $v_o$, we also randomly sample a set of pseudo views $\Vec{v}_p = \{v_{p_i}\}_{i=0}^{M}$ in every $k$ training iterations. The position of the pseudo views is interpolated between the current training view and its adjacent views. And the orientation of the pseudo views is rotated from the training views within the range of $[-\delta, \delta]$ along the z-axis (yaw angle).

The loss function to regularize the 3D GS training is similar to the sample loss from \cite{wu2023reconfusion}, as it has been proved to perform better than score distillation sampling (SDS)\cite{wu2023reconfusion,poole2022dreamfusion}. The pseudo views are also rendered via differentiable splatting, denoted as $\Tilde{\mathbf{I}}_{p}(\Phi_{p}, \Vec{v}_p) \in \mathbb{R}^{M,3,H,W}$, where $\Phi_p=\{\Vec{\phi}_i\}_{i \in N_p}$ is the parameters of Gaussian models corresponding to the pseudo views $\Vec{v}_p$. Then the rendered images are fed into our Diffusion Model to generate the guidance image $\mathbf{I}_g$ in an image-to-image manner. The rendered pseudo views are encoded into latents and are added noise with a randomly selected noisy level $t$ between the max noisy level $t_{\max}$ and the min noisy level $t_{\min}$. The noisy latent is denoised from the selected noisy level to $t_{\min}$ and then decoded to obtain the guidance images $\mathbf{I}_g$. The loss function between the guidance images and rendered images from pseudo views is formulated as:
\begin{equation}
    L_\text{pseudo}(\Phi_p) = \mathbb{E}_{\Vec{v}_p} [ ||\mathbf{I}_g, \Tilde{\mathbf{I}}_p||_1 + \lambda_\text{p-lpips}L_\text{lpips}(\mathbf{I}_g, \Tilde{\mathbf{I}}_p) + \lambda_\text{p-depth}||\Vec{D}_p, \Tilde{\Vec{D}}_p||_1]
\end{equation}
The overall loss function is as follows:
\begin{equation}
    L(\Phi) = L_\text{recon}(\Phi_o) + \lambda_\text{pseudo}L_\text{pseudo}(\Phi_p), \quad \Phi = \text{Union}\{\Phi_o, \Phi_p\}
\end{equation}
where $\Phi$ denotes the union of the Gaussian Models' parameters corresponding to the training view and the pseudo views. Because the pseudo views are sampled close to training views, there exists a considerable overlap among their corresponding Gaussian Models. This allows these Models to learn from multi-views simultaneously, preventing them from getting stuck in local minima. Meanwhile, the plausible images from the pseudo views recovered by our fine-tuned Diffusion Model afford the Gaussian Models a more comprehensive observation of the scene, enhancing the capability for free-view rendering.
% TODO: Fig
% 
% \subsubsection{Training}

\section{Experiments}
\label{sec:experiments}
\subsection{Implementation Details }
% \subsection{Implementation Details}
% - structure of our modified Diffusion Model
% - Data pre-processing when fine-tuning Diffusion Model
% - 
%  The text prompt is chosen from some templates \texttt{"A \{image, photo, ...\} of a \{street, driving, ...\} scene"} to guide the start of the fine-tuning.
\subsubsection{Diffusion Model} Our Diffusion Model is fine-tuned based on Stable Diffusion 1.5\cite{rombach2022sd}. The additional CLIP Image Encoder is taken from clip-vit-B-32\cite{radford2021clip}. It takes images with size $224 \times 224 \times 3$ as input, and its hidden state dimension is $1 \times 50 \times 768$, where 50 is the token length. The first token is the \textit{cls} token and the last 49 tokens are the patch tokens. The Depth Encoder is pre-trained in the mechanism proposed from \cite{huang2023clip2point}. Its structure is identical to the CLIP Image Encoder. The Depth ControlNet is initialized as a trainable copy of the fine-tuned U-Net encoder from training stage one. In practice, we fine-tuned the Diffusion Model's U-Net with 625,000 iterations in the first stage and trained the ControlNet with 125,000 steps in the second stage. Both are on 4 32G V100 GPUs with batch size 4.

\subsubsection{3D Gaussian Splatting}. For initializing the Gaussian models of the scene, we only utilize the LiDAR point clouds which is voxelized-downsampled with a voxel size 0.5. No SfM (Structure from Motion) point clouds are used in our approach as we intend to avoid the impact from the prior provided by SfM point clouds and to ensure the exclusive use of data collected by vehicles. We trained the 3DGS model for 50,000 iterations with the learning rate decreasing from $1.6 \times 10^{-4}$ to $1.6 \times 10^{-6}$, which is identical to the original 3DGS.

\subsection{Experiment Setup}
% - Datasets: Kitti, Kitti360
% - Baseline: 3DGS; Other methods:（ZipNeRF）
% - Evaluation: 
%   - Test Views: PSNR SSIM LPIPS
%   — Novel Views: FID, NIQE, BRISQUE
\subsubsection{Datasets} We evaluate our method on two widely-used autonomous driving datasets KITTI \cite{geiger2015kitti} and KITTI-360 \cite{liao2022kitti}. Both datasets provide forward-looking images and compact LiDAR point clouds. To prove the generalization ability, we only fine-tuned our Diffusion Model on about 12,000 images randomly selected from KITTI-360 datasets\cite{liao2022kitti}. All images from KITTI datasets\cite{geiger2015kitti} are not seen during the fine-tuning. When training the 3DGS model, only monocular images are used.

\subsubsection{Competitors} Since our method is built upon 3DGS framework, we establish it as our baseline for comparison. To ensure a fair comparison, we also implement the depth loss for 3DGS, and keep its hyper-parameters identical to our method. For comparative analysis, we also select Zip-NeRF\cite{barron2023zipnerf}, which is the SOTA method in rendering qualities, and MARS\cite{wu2023mars}, which is the SOTA method in street view synthesis, as our competitors. For Zip-NeRF, we use the PyTorch implementation with nerfstudio\cite{nerfstudio}.

\subsection{Comparison Results}

\subsubsection{Evaluation on Test Views}
For evaluating the rendering quality on the test views, we adopt three commonly used metrics PSNR, SSIM and LPIPS \cite{zhang2018lpips}. \Cref{tab:testviews_kitti,tab:testviews_kitti360} respectively present the quantitative comparison of our method again 3DGS\cite{3dgs}, Zip-NeRF\cite{barron2023zipnerf} and MARS\cite{wu2023mars} on KITTI\cite{geiger2015kitti} and KITTI-360\cite{liao2022kitti} datasets. In line with MARS\cite{wu2023mars}, we evaluate our method on KITTI with 75\%, 50\% and 25\% training/testing splits. For KITTI-360 dataset, we follow the setting of its official $50\%$ drop-rate NVS benchmark\cite{liao2022kitti}.

Across all conducted experiments, our method substantially outperforms our baseline method 3DGS. Under the KITTI-75\% setting, our results are inferior to those of MARS. This is primarily due to the uniform modeling of static background and dynamic objects in our method. This limitation can be similarly observed in Zip-NeRF\cite{barron2023zipnerf}. However, at the sparse-view input setting in KITTI-25\%, our method surpasses all the competitors. This improvement is attributed to the additional information of the unobserved view provided by the Diffusion Model. On KITTI-360 dataset, our method achieve the best performance among all competitors.

\begin{table}[htb]
    \caption{\textbf{Quantitative results on the KITTI dataset\cite{geiger2015kitti}.} *The results of MARS\cite{wu2023mars} are taken from their original paper.
  }
    \label{tab:testviews_kitti}
    \centering
    \begin{tabular}{@{}lccccccccc@{}}   
        \toprule
        & \multicolumn{3}{c}{KITTI - 75\%} & \multicolumn{3}{c}{KITTI - 50\%} & \multicolumn{3}{c}{KITTI - 25\%} \\
        \cmidrule(r){2-4} \cmidrule(lr){5-7} \cmidrule(l){8-10}
        & PSNR↑ & SSIM↑ & LPIPS↓ & PSNR↑ & SSIM↑ & LPIPS↓ & PSNR↑ & SSIM↑ & LPIPS↓ \\ 
        \midrule
        3DGS \cite{3dgs} & 22.40 & 0.805 & 0.183 & 21.10 & 0.752 & 0.187 & 19.98 & 0.741 &  0.180\\
        Zip-NeRF \cite{barron2023zipnerf} & 22.51 & 0.827 & 0.173 & 21.23 & 0.789 & 0.179 & 20.30 & \cellcolor{orange!20}0.766 & 0.185 \\
        MARS* \cite{wu2023mars} & \cellcolor{red!20}24.23 & \cellcolor{red!20}0.845 & \cellcolor{orange!20}0.160 & \cellcolor{red!20}24.00 & \cellcolor{red!20}0.801 & \cellcolor{orange!20}0.164 & \cellcolor{orange!20}23.23 & 0.756 & \cellcolor{orange!20}0.177 \\
        Ours & \cellcolor{orange!20}23.85 & \cellcolor{orange!20}0.837 & \cellcolor{red!20}0.154 & \cellcolor{orange!20}23.62 & \cellcolor{orange!20}0.799 & \cellcolor{red!20}0.158 & \cellcolor{red!20}23.44 & \cellcolor{red!20}0.793 & \cellcolor{red!20}0.167 \\
        \bottomrule
    \end{tabular}
\end{table}

\vspace{-1cm}
\begin{table}[htb]
    \caption{\textbf{Quantitative results on the KITTI-360 dataset\cite{liao2022kitti}.} *The results of MARS\cite{wu2023mars} are taken from the leaderboard of KITTI-360 NVS benchmark\cite{liao2022kitti}.
      }
    \label{tab:testviews_kitti360}
    \centering
    \begin{tabular}{@{}lccc@{}}
        \toprule
        & \multicolumn{3}{c}{KITTI-360 - 50\%}\\
        \cmidrule(r){2-4} & \makebox[0.15\textwidth]{PSNR↑} & \makebox[0.15\textwidth]{SSIM↑} & \makebox[0.15\textwidth]{LPIPS↓} \\ 
        \midrule
        3DGS \cite{3dgs} & 22.78 & 0.793 & 0.176 \\
        Zip-NeRF \cite{barron2023zipnerf} & 22.86 & 0.802 & \cellcolor{orange!20}0.167\\
        MARS* \cite{wu2023mars} & \cellcolor{orange!20}23.09 & \cellcolor{red!20}0.857 & 0.174 \\
        Ours & \cellcolor{red!20}23.81 & \cellcolor{orange!20}0.832 & \cellcolor{red!20}0.155 \\
        \bottomrule
    \end{tabular}
\end{table}

\subsubsection{Evaluation on Novel Views}
To evaluate each model's capability for free-view rendering, we also select novel views that are distant from the training and testing view for evaluation. These novel viewpoints are created by interpolating the position of the training/testing views and adding some perturbations. Their rotation is adjusted by offsetting angle $\pm \delta, \pm 2\delta$ along the z-axis, where $\delta$ is randomly chosen from $[15^\circ, 30^\circ]$. Given the absence of corresponding ground-truth images for these novel views, we adopt a no-reference image quality assessment method BRISQUE\cite{mittal2012brisque} to quantitatively measure the image quality, and the FID (Fréchet Inception Distance) score \cite{heusel2017gans} to measure the difference of distribution between the rendered novels views and the training images, while qualitatively comparing rendering cases. 

\begin{figure}[htb]
  \centering
  \includegraphics[width=0.94\linewidth]{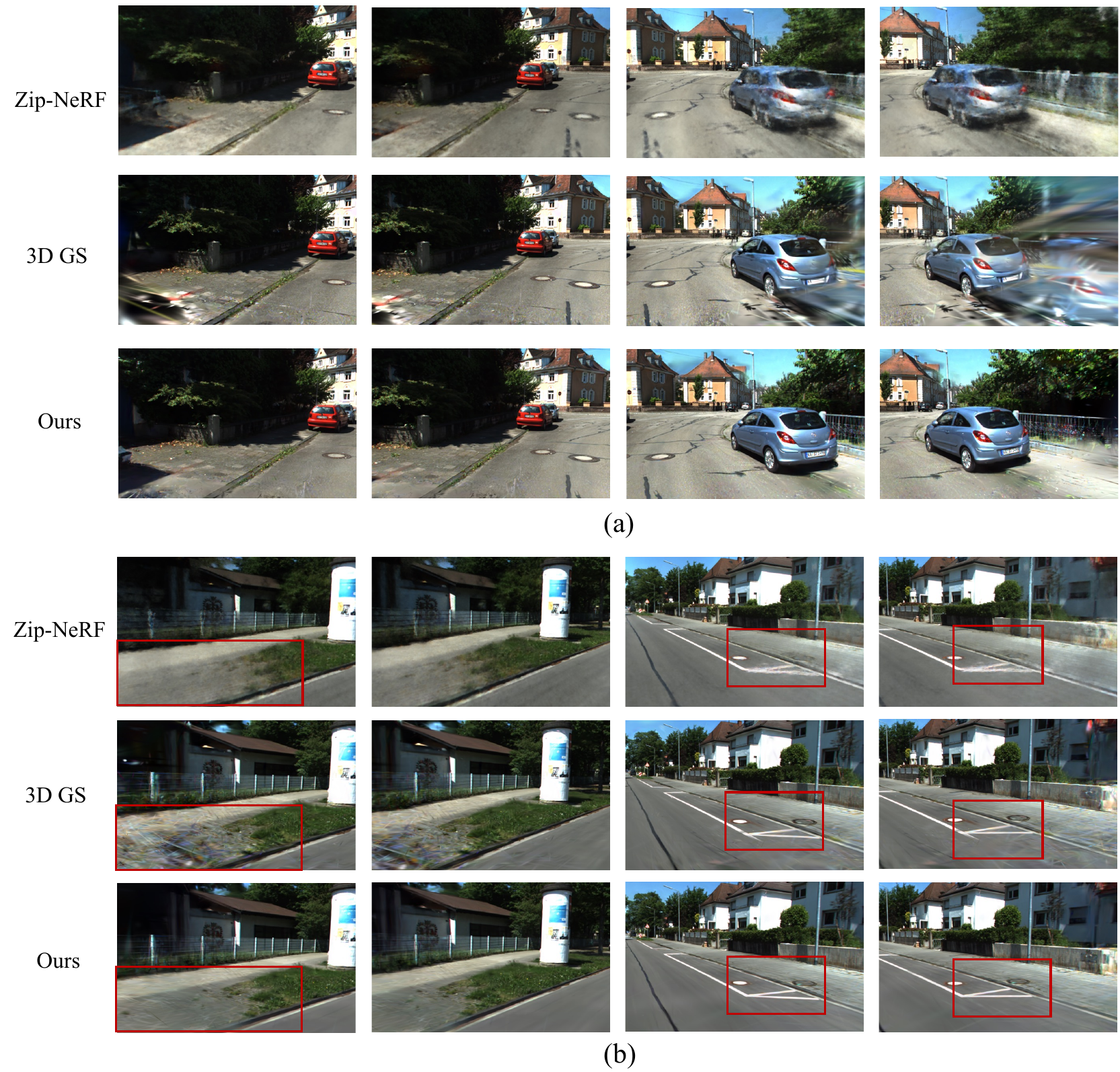}
  \caption{\textbf{Qualitative comparisons of novel views rendering on the KITTI-360\cite{liao2022kitti} dataset.} ZipNeRF\cite{barron2023zipnerf} and 3DGS\cite{3dgs} produce artifacts of the blue vehicle in (a) and blurry lane markings in (b), while our method preserves high rendering quality. Our method also fix the hole on the road surface generated by 3DGS\cite{3dgs}.
  }
  \label{fig:novelviews_kitti360}
\end{figure}

\begin{figure}[htb]
  \centering
  \includegraphics[width=0.95\linewidth]{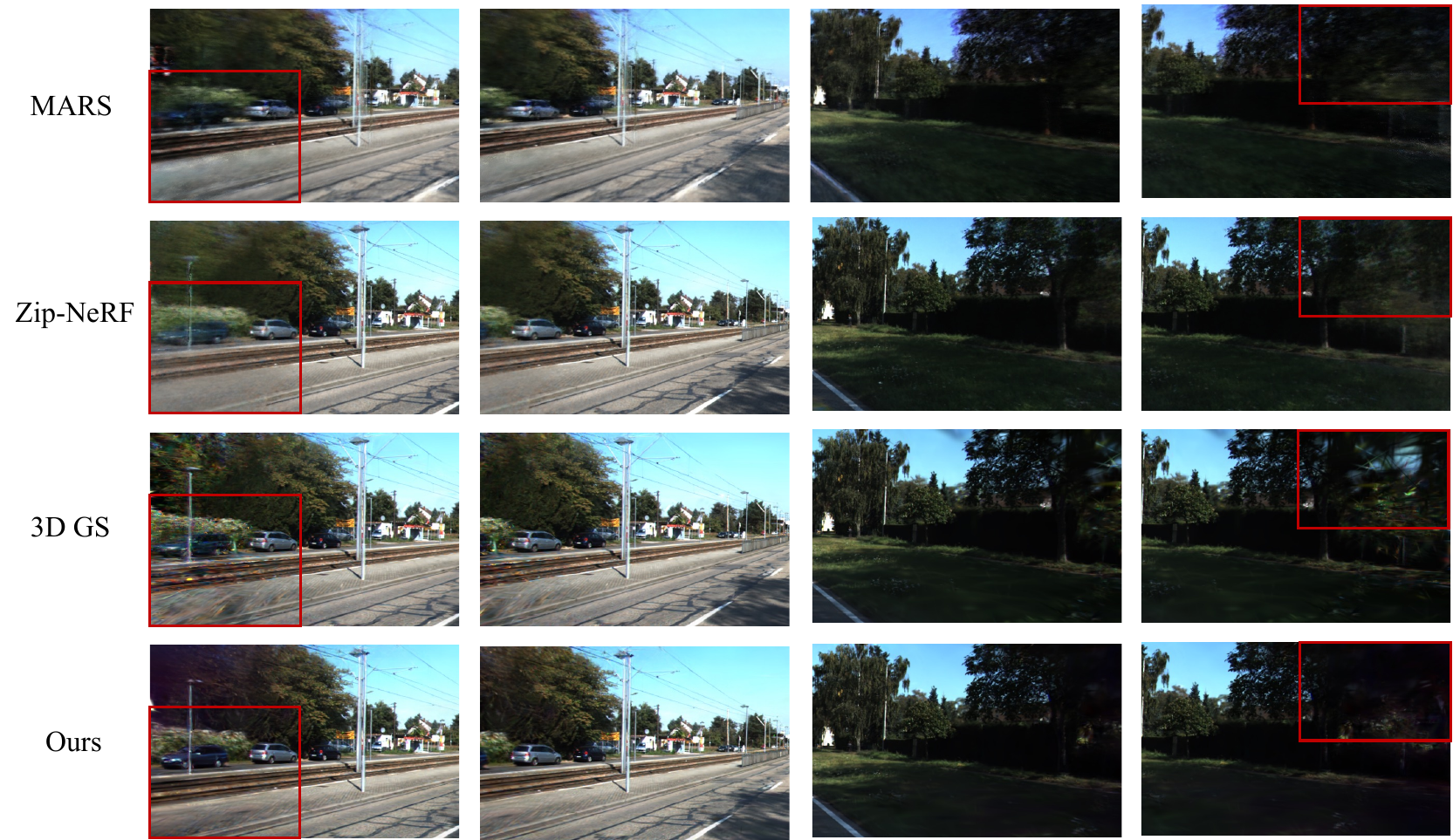}
  \caption{\textbf{Qualitative comparisons of novel views rendering on the KITTI\cite{geiger2015kitti} dataset.}
  }
  \label{fig:novelviews_kitti}
\end{figure}

As indicated in \Cref{tab:novelviews}, our method achieves the lowest BRISQUE score among all methods, suggesting that our rendered images in novel views preserve high quality with less noise and blur. Our method also exhibits the lowest FID score, reflecting a better alignment with the original images of street scenes. \cref{fig:novelviews_kitti,fig:novelviews_kitti360} illustrates the qualitative results among the methods on KITTI\cite{geiger2015kitti} and KITTI-360\cite{liao2022kitti}. It can be clearly seen in \cref{fig:novelviews_kitti360}(a) that 3DGS and Zip-NeRF both produce severe artifacts of the blue vehicle. This is because they are trained predominantly by the rear view of the vehicle without observing it from other directions. Our method successfully mitigates this issue by incorporating the augmented views generated by the Diffusion Model. \cref{fig:novelviews_kitti360}(b) and \cref{fig:novelviews_kitti} further demonstrate that our method enhances the rendering quality, resulting in smoother road surface, more distinct lane markings, and clearer vehicles in the distance.

\begin{table}[htb]
    \centering
    \caption{\textbf{Image quality evaluation on novel views.}}
    \label{tab:novelviews}
    \centering
    \begin{tabular}{@{}lcc@{}}
        \toprule
         & \makebox[0.15\textwidth][c]{BRISQUE↓} & \makebox[0.15\textwidth][c]{FID↓} \\ 
        \midrule
        GT & 20.71  &  —— \\
        \midrule
        3DGS \cite{3dgs} & 32.64 &  \cellcolor{orange!20}81.39\\
        Zip-NeRF \cite{barron2023zipnerf} & \cellcolor{orange!20}27.02 & 92.72 \\
        Ours & \cellcolor{red!20}24.53 & \cellcolor{red!20}73.36 \\
        \bottomrule
    \end{tabular}
\end{table}

\begin{figure}[!h]
  \centering
  \includegraphics[width=0.95\linewidth]{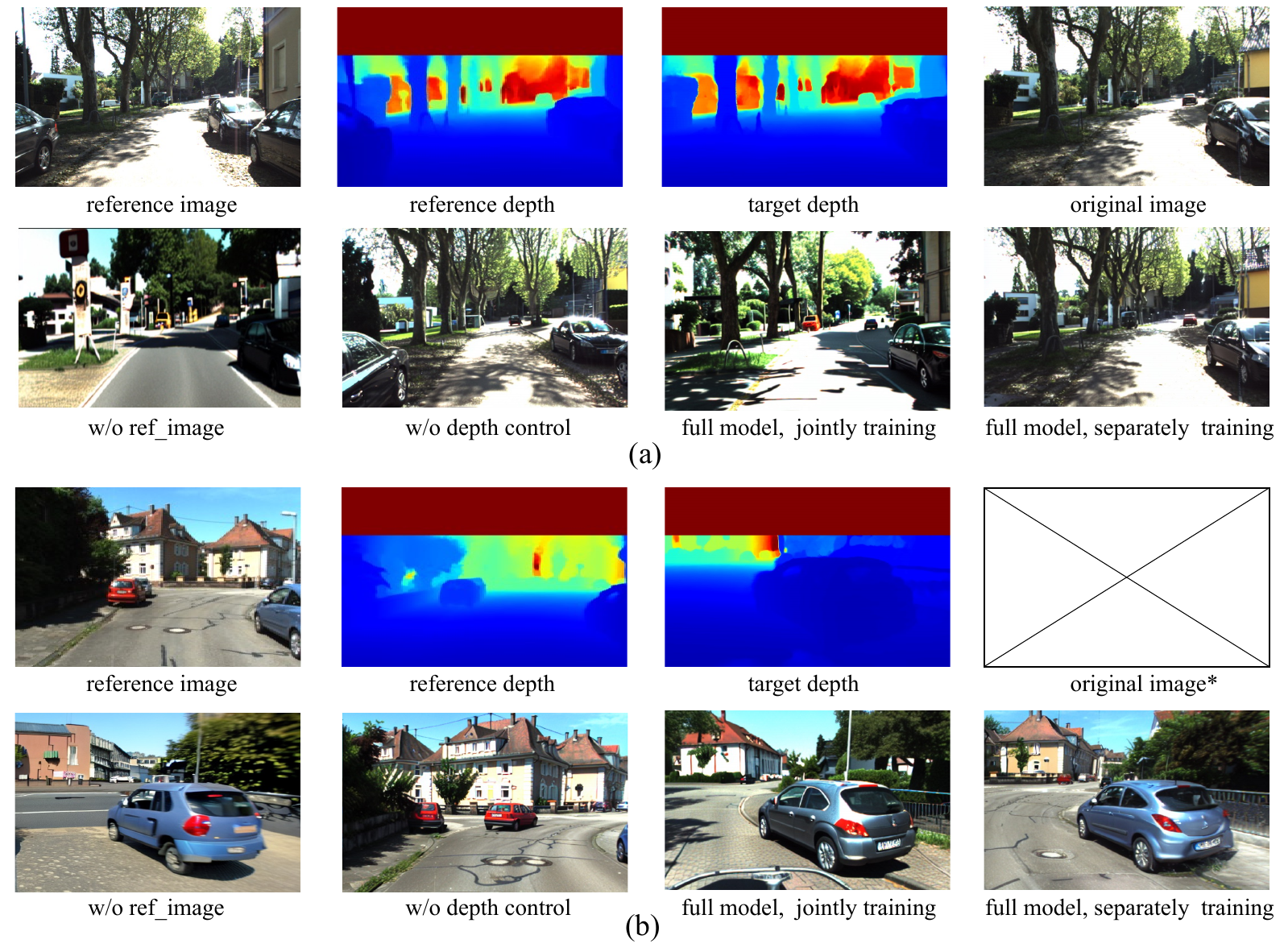}
  \caption{\textbf{Qualitative ablation results on different conditions and different fine-tuning schemes of Diffusion Model\cite{rombach2022sd}.} *Target view in (b) is a novel view thus its original image is left blank.
  }
  \label{fig:ablation-1}
\end{figure}

\subsection{Ablation Studies}
% Diffusion Model: Ablation on ref image & depth control
% 3dgs + diffusion: Ablation on different losses
We conduct ablation experiments on the two main processes within our method: the fine-tuning of the Diffusion Model and the 3DGS training process. 

When fine-tuning the Diffusion Model, our method utilizes reference images and depth as conditions and conducts a two-phase training strategy. The influences of different condition signals and different training schemes are evaluated, as depicted in \cref{fig:ablation-1}. The upper row of \cref{fig:ablation-1}(a) and (b) shows the reference image, the target image, and the depth map for both the reference view and target view. The target view in \cref{fig:ablation-1}(b) is a novel view, thus its original image is left blank. When solely utilizing a depth ControlNet while not considering reference images, the semantic information of the generated image is governed by the text prompt, resulting in a high diversity and significant deviations from the original images. Conversely, introducing reference images as the sole conditional input without depth guidance allows the Diffusion Model to assimilate scene semantics without understanding the precise locational context of scene objects. Such as in \cref{fig:ablation-1}, without the depth information from the reference image, the model recognizes that there is a red car in the scene but misplaces it. Fine-tuning the U-Net and training the ControlNet simultaneously would result in less authentic outputs. This is due to, in such training scheme, the ControlNet has to be initialized by the weights of the original Stable Diffusion 1.5 other than the fine-tuned ones.

We also ablate the loss functions that regularize the training of pseudo views. As illustrated in \cref{tab:ablation_loss}, each loss validates its efficacy. However, it was observed that introducing $L_1$ loss for pseudo views marginally reduced the PSNR of the test views. This can be attributed to the inherent minor differences between the images produced by the Diffusion Model and the actual scenes, which become apparent upon pixel-level comparison.

\begin{table}[htb]
    \centering
    \caption{\bf{Quantitative ablation result on each module in 3DGS training.}}
    \label{tab:ablation_loss}
    \centering
    \begin{tabular}{@{}lccc@{}}
        \toprule
         & \makebox[0.12\textwidth][c]{PSNR↑} & \makebox[0.12\textwidth][c]{SSIM↑} & \makebox[0.12\textwidth][c]{LPIPS↓}\\ 
        \midrule
        Complete model & 23.81 & 0.832 & 0.155   \\
        \midrule
        w/o $L_{\text{LPIPS} \_\text{pseudo}}$   & 23.11  & 0.818 & 0.159 \\
        w/o $L_{1\_\text{pseudo}}$  & \cellcolor{blue!10}23.83  &0.820 &  0.163 \\
        w/o $L_{\text{depth}\_\text{pseudo}}$  & 23.29 & 0.801 & 0.169\\
        w/o pseudo views (3DGS) & 22.78 & 0.793 & 0.176 \\
        \bottomrule
    \end{tabular}
\end{table}

\section{Discussion and Conclusion}
\label{sec:conclusion}

% \begin{wrapfigure}{r}{0.45\linewidth}
% \begin{figure}[t]
%   \centering
%   \includegraphics[width=0.6\linewidth]{figs/time2.pdf}
%   \caption{ \textbf{Relationship between the number of sampled pseudo views per iteration, training time (in Iterations/seconds), and rendering time (in FPS).} When pseudo views are introduced the training speed decreases significantly, while the inference speed is not impacted.}
%   \label{fig:time}
% % \end{wrapfigure}
% \end{figure}
\begin{figure}[tb]
  \centering
  \begin{subfigure}{0.48\linewidth}
    \includegraphics[width=\linewidth]{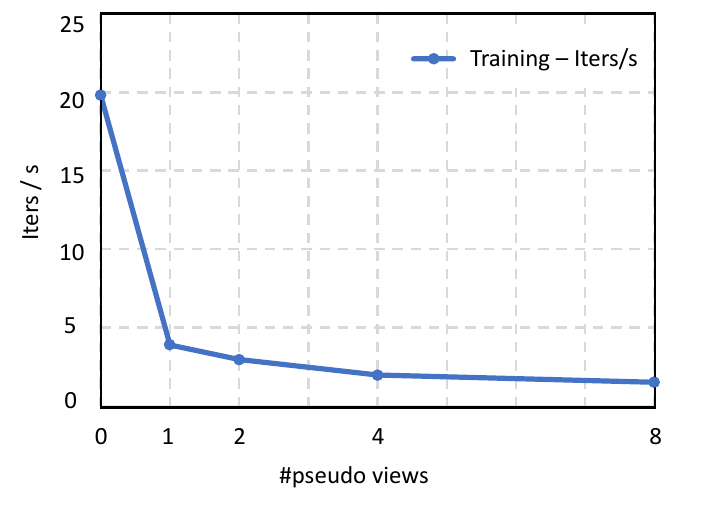}
    \caption{}
    \label{fig:time-1}
  \end{subfigure}
  \hfill
  \begin{subfigure}{0.48\linewidth}
    \includegraphics[width=\linewidth]{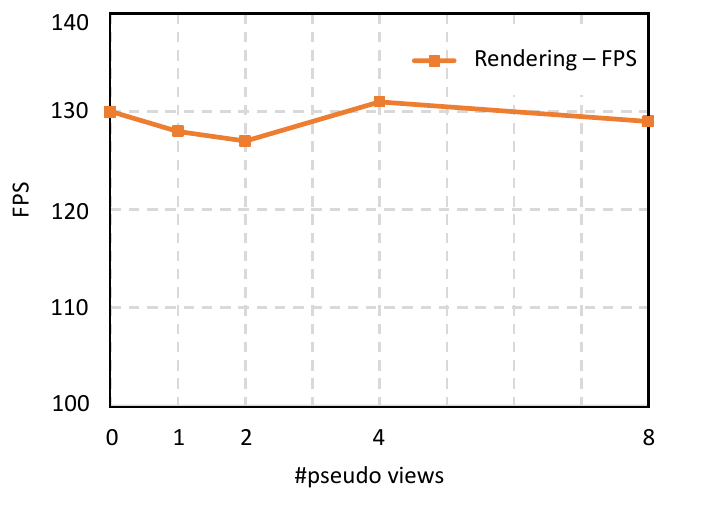}
    \caption{}
    \label{fig:time-2}
  \end{subfigure}
  \caption{\textbf{Relationship between the number of sampled pseudo views per iteration and (a). training speed (in iterations per second), (b). rendering speed (in frames per second (FPS)).} When pseudo views are introduced the training speed decreases significantly, while the inference speed is not impacted.}
  \label{fig:time}
\end{figure}

\vspace{-1cm}

\subsubsection{Limitation} The integration of Diffusion Model into 3DGS introduces a notable limitation: longer training time. It is primarily caused by the time-consuming denoising operation of Diffusion Model. \cref{fig:time-1} shows the correlation between the number of sample pseudo views and the training speed. A substantial decrease in training speed can be observed with an increment from 0 pseudo views (standard 3DGS) to 1. Since our method does not affect the real-time inference ability of 3DGS, as illustrated in \cref{fig:time-2}, and yields proved render quality, we temporarily accept the training time and leave improving the training efficiency as our future work.

In conclusion, we present a method aimed at enhancing the capability of free-viewpoint rendering within autonomous driving scenarios. While certain limitations persist, our method has shown proficiency in maintaining high-quality renderings from novel viewpoints, with considerable efficiency in rendering. This allows our method to offer a broader perspective within autonomous driving simulations, enabling the simulation of potentially hazardous corner cases, and thus enhancing the overall safety and reliability of autonomous driving systems.

% \clearpage  % TODO REVIEW/FINAL: This \clearpage needs to be removed from both review and camera-ready versions.

\bibliographystyle{splncs04}
\bibliography{main}

\begin{thebibliography}{10}
\providecommand{\url}[1]{\texttt{#1}}
\providecommand{\urlprefix}{URL }
\providecommand{\doi}[1]{https://doi.org/#1}

\bibitem{barron2023zipnerf}
Barron, J.T., Mildenhall, B., Verbin, D., Srinivasan, P.P., Hedman, P.: Zip-nerf: Anti-aliased grid-based neural radiance fields. ICCV  (2023)

\bibitem{chan2023genvs}
Chan, E.R., Nagano, K., Chan, M.A., Bergman, A.W., Park, J.J., Levy, A., Aittala, M., De~Mello, S., Karras, T., Wetzstein, G.: Generative novel view synthesis with 3d-aware diffusion models. arXiv preprint arXiv:2304.02602  (2023)

\bibitem{chang2015shapenet}
Chang, A.X., Funkhouser, T., Guibas, L., Hanrahan, P., Huang, Q., Li, Z., Savarese, S., Savva, M., Song, S., Su, H., et~al.: Shapenet: An information-rich 3d model repository. arXiv preprint arXiv:1512.03012  (2015)

\bibitem{chen2023periodic}
Chen, Y., Gu, C., Jiang, J., Zhu, X., Zhang, L.: Periodic vibration gaussian: Dynamic urban scene reconstruction and real-time rendering. arXiv preprint arXiv:2311.18561  (2023)

\bibitem{deitke2023objaverse}
Deitke, M., Schwenk, D., Salvador, J., Weihs, L., Michel, O., VanderBilt, E., Schmidt, L., Ehsani, K., Kembhavi, A., Farhadi, A.: Objaverse: A universe of annotated 3d objects. In: Proceedings of the IEEE/CVF Conference on Computer Vision and Pattern Recognition. pp. 13142--13153 (2023)

\bibitem{deng2022dsnerf}
Deng, K., Liu, A., Zhu, J.Y., Ramanan, D.: Depth-supervised nerf: Fewer views and faster training for free. In: Proceedings of the IEEE/CVF Conference on Computer Vision and Pattern Recognition. pp. 12882--12891 (2022)

\bibitem{carla}
Dosovitskiy, A., Ros, G., Codevilla, F., Lopez, A., Koltun, V.: Carla: An open urban driving simulator. In: Conference on robot learning. pp. 1--16. PMLR (2017)

\bibitem{geiger2015kitti}
Geiger, A., Lenz, P., Stiller, C., Urtasun, R.: The kitti vision benchmark suite. URL http://www. cvlibs. net/datasets/kitti  \textbf{2}(5) (2015)

\bibitem{guo2023streetsurf}
Guo, J., Deng, N., Li, X., Bai, Y., Shi, B., Wang, C., Ding, C., Wang, D., Li, Y.: Streetsurf: Extending multi-view implicit surface reconstruction to street views. arXiv preprint arXiv:2306.04988  (2023)

\bibitem{heusel2017gans}
Heusel, M., Ramsauer, H., Unterthiner, T., Nessler, B., Hochreiter, S.: Gans trained by a two time-scale update rule converge to a local nash equilibrium. Advances in neural information processing systems  \textbf{30} (2017)

\bibitem{huang2023clip2point}
Huang, T., Dong, B., Yang, Y., Huang, X., Lau, R.W., Ouyang, W., Zuo, W.: Clip2point: Transfer clip to point cloud classification with image-depth pre-training. In: Proceedings of the IEEE/CVF International Conference on Computer Vision. pp. 22157--22167 (2023)

\bibitem{3dgs}
Kerbl, B., Kopanas, G., Leimk{\"u}hler, T., Drettakis, G.: 3d gaussian splatting for real-time radiance field rendering. ACM Transactions on Graphics  \textbf{42}(4) (2023)

\bibitem{kwak2023geconerf}
Kwak, M.S., Song, J., Kim, S.: Geconerf: Few-shot neural radiance fields via geometric consistency. arXiv preprint arXiv:2301.10941  (2023)

\bibitem{liao2022kitti}
Liao, Y., Xie, J., Geiger, A.: Kitti-360: A novel dataset and benchmarks for urban scene understanding in 2d and 3d. IEEE Transactions on Pattern Analysis and Machine Intelligence  \textbf{45}(3),  3292--3310 (2022)

\bibitem{liu2023real}
Liu, J.Y., Chen, Y., Yang, Z., Wang, J., Manivasagam, S., Urtasun, R.: Real-time neural rasterization for large scenes. In: Proceedings of the IEEE/CVF International Conference on Computer Vision. pp. 8416--8427 (2023)

\bibitem{liu2023zero}
Liu, R., Wu, R., Van~Hoorick, B., Tokmakov, P., Zakharov, S., Vondrick, C.: Zero-1-to-3: Zero-shot one image to 3d object. In: Proceedings of the IEEE/CVF International Conference on Computer Vision. pp. 9298--9309 (2023)

\bibitem{lu2023dnmp}
Lu, F., Xu, Y., Chen, G., Li, H., Lin, K.Y., Jiang, C.: Urban radiance field representation with deformable neural mesh primitives. In: Proceedings of the IEEE/CVF International Conference on Computer Vision. pp. 465--476 (2023)

\bibitem{mildenhall2021nerf}
Mildenhall, B., Srinivasan, P.P., Tancik, M., Barron, J.T., Ramamoorthi, R., Ng, R.: Nerf: Representing scenes as neural radiance fields for view synthesis. Communications of the ACM  \textbf{65}(1),  99--106 (2021)

\bibitem{mittal2012brisque}
Mittal, A., Moorthy, A.K., Bovik, A.C.: No-reference image quality assessment in the spatial domain. IEEE Transactions on image processing  \textbf{21}(12),  4695--4708 (2012)

\bibitem{nsg}
Ost, J., Mannan, F., Thuerey, N., Knodt, J., Heide, F.: Neural scene graphs for dynamic scenes. In: Proceedings of the IEEE/CVF Conference on Computer Vision and Pattern Recognition. pp. 2856--2865 (2021)

\bibitem{poole2022dreamfusion}
Poole, B., Jain, A., Barron, J.T., Mildenhall, B.: Dreamfusion: Text-to-3d using 2d diffusion. In: The Eleventh International Conference on Learning Representations (2022)

\bibitem{radford2021clip}
Radford, A., Kim, J.W., Hallacy, C., Ramesh, A., Goh, G., Agarwal, S., Sastry, G., Askell, A., Mishkin, P., Clark, J., et~al.: Learning transferable visual models from natural language supervision. In: International conference on machine learning. pp. 8748--8763. PMLR (2021)

\bibitem{reizenstein2021co3d}
Reizenstein, J., Shapovalov, R., Henzler, P., Sbordone, L., Labatut, P., Novotny, D.: Common objects in 3d: Large-scale learning and evaluation of real-life 3d category reconstruction. In: Proceedings of the IEEE/CVF International Conference on Computer Vision. pp. 10901--10911 (2021)

\bibitem{rematas2022urf}
Rematas, K., Liu, A., Srinivasan, P.P., Barron, J.T., Tagliasacchi, A., Funkhouser, T., Ferrari, V.: Urban radiance fields. In: Proceedings of the IEEE/CVF Conference on Computer Vision and Pattern Recognition. pp. 12932--12942 (2022)

\bibitem{roessle2022ddpnerf}
Roessle, B., Barron, J.T., Mildenhall, B., Srinivasan, P.P., Nie{\ss}ner, M.: Dense depth priors for neural radiance fields from sparse input views. In: Proceedings of the IEEE/CVF Conference on Computer Vision and Pattern Recognition. pp. 12892--12901 (2022)

\bibitem{roessle2023ganerf}
Roessle, B., M{\"u}ller, N., Porzi, L., Bul{\`o}, S.R., Kontschieder, P., Nie{\ss}ner, M.: Ganerf: Leveraging discriminators to optimize neural radiance fields. arXiv preprint arXiv:2306.06044  (2023)

\bibitem{rombach2022sd}
Rombach, R., Blattmann, A., Lorenz, D., Esser, P., Ommer, B.: High-resolution image synthesis with latent diffusion models. In: Proceedings of the IEEE/CVF conference on computer vision and pattern recognition. pp. 10684--10695 (2022)

\bibitem{rong2020lgsvl}
Rong, G., Shin, B.H., Tabatabaee, H., Lu, Q., Lemke, S., Mo{\v{z}}eiko, M., Boise, E., Uhm, G., Gerow, M., Mehta, S., et~al.: Lgsvl simulator: A high fidelity simulator for autonomous driving. In: 2020 IEEE 23rd International conference on intelligent transportation systems (ITSC). pp.~1--6. IEEE (2020)

\bibitem{sargent2023zeronvs}
Sargent, K., Li, Z., Shah, T., Herrmann, C., Yu, H.X., Zhang, Y., Chan, E.R., Lagun, D., Fei-Fei, L., Sun, D., et~al.: Zeronvs: Zero-shot 360-degree view synthesis from a single real image. arXiv preprint arXiv:2310.17994  (2023)

\bibitem{shah2018airsim}
Shah, S., Dey, D., Lovett, C., Kapoor, A.: Airsim: High-fidelity visual and physical simulation for autonomous vehicles. In: Field and Service Robotics: Results of the 11th International Conference. pp. 621--635. Springer (2018)

\bibitem{shi2023zero123++}
Shi, R., Chen, H., Zhang, Z., Liu, M., Xu, C., Wei, X., Chen, L., Zeng, C., Su, H.: Zero123++: a single image to consistent multi-view diffusion base model. arXiv preprint arXiv:2310.15110  (2023)

\bibitem{somraj2023VipNeRF}
Somraj, N., Soundararajan, R.: {ViP-NeRF}: Visibility prior for sparse input neural radiance fields  (August 2023). \doi{10.1145/3588432.3591539}

\bibitem{tancik2022block}
Tancik, M., Casser, V., Yan, X., Pradhan, S., Mildenhall, B., Srinivasan, P.P., Barron, J.T., Kretzschmar, H.: Block-nerf: Scalable large scene neural view synthesis. In: Proceedings of the IEEE/CVF Conference on Computer Vision and Pattern Recognition. pp. 8248--8258 (2022)

\bibitem{nerfstudio}
Tancik, M., Weber, E., Ng, E., Li, R., Yi, B., Kerr, J., Wang, T., Kristoffersen, A., Austin, J., Salahi, K., Ahuja, A., McAllister, D., Kanazawa, A.: Nerfstudio: A modular framework for neural radiance field development. In: ACM SIGGRAPH 2023 Conference Proceedings. SIGGRAPH '23 (2023)

\bibitem{tonderski2023neurad}
Tonderski, A., Lindstr{\"o}m, C., Hess, G., Ljungbergh, W., Svensson, L., Petersson, C.: Neurad: Neural rendering for autonomous driving. arXiv preprint arXiv:2311.15260  (2023)

\bibitem{turki2022mega}
Turki, H., Ramanan, D., Satyanarayanan, M.: Mega-nerf: Scalable construction of large-scale nerfs for virtual fly-throughs. In: Proceedings of the IEEE/CVF Conference on Computer Vision and Pattern Recognition. pp. 12922--12931 (2022)

\bibitem{turki2023suds}
Turki, H., Zhang, J.Y., Ferroni, F., Ramanan, D.: Suds: Scalable urban dynamic scenes. In: Proceedings of the IEEE/CVF Conference on Computer Vision and Pattern Recognition. pp. 12375--12385 (2023)

\bibitem{verbin2022refnerf}
Verbin, D., Hedman, P., Mildenhall, B., Zickler, T., Barron, J.T., Srinivasan, P.P.: Ref-nerf: Structured view-dependent appearance for neural radiance fields. In: 2022 IEEE/CVF Conference on Computer Vision and Pattern Recognition (CVPR). pp. 5481--5490. IEEE (2022)

\bibitem{wang2023sparsenerf}
Wang, G., Chen, Z., Loy, C.C., Liu, Z.: Sparsenerf: Distilling depth ranking for few-shot novel view synthesis. arXiv preprint arXiv:2303.16196  (2023)

\bibitem{wu2023reconfusion}
Wu, R., Mildenhall, B., Henzler, P., Park, K., Gao, R., Watson, D., Srinivasan, P.P., Verbin, D., Barron, J.T., Poole, B., Holynski, A.: Reconfusion: 3d reconstruction with diffusion priors. arXiv  (2023)

\bibitem{wu2023mars}
Wu, Z., Liu, T., Luo, L., Zhong, Z., Chen, J., Xiao, H., Hou, C., Lou, H., Chen, Y., Yang, R., et~al.: Mars: An instance-aware, modular and realistic simulator for autonomous driving. In: CAAI International Conference on Artificial Intelligence. pp. 3--15. Springer (2023)

\bibitem{wynn2023diffusionerf}
Wynn, J., Turmukhambetov, D.: Diffusionerf: Regularizing neural radiance fields with denoising diffusion models. In: Proceedings of the IEEE/CVF Conference on Computer Vision and Pattern Recognition. pp. 4180--4189 (2023)

\bibitem{snerf}
Xie, Z., Zhang, J., Li, W., Zhang, F., Zhang, L.: S-nerf: Neural radiance fields for street views. arXiv preprint arXiv:2303.00749  (2023)

\bibitem{xiong2023sparsegs}
Xiong, H., Muttukuru, S., Upadhyay, R., Chari, P., Kadambi, A.: Sparsegs: Real-time 360 $\{$$\backslash$deg$\}$ sparse view synthesis using gaussian splatting. arXiv preprint arXiv:2312.00206  (2023)

\bibitem{yan2024street}
Yan, Y., Lin, H., Zhou, C., Wang, W., Sun, H., Zhan, K., Lang, X., Zhou, X., Peng, S.: Street gaussians for modeling dynamic urban scenes. arXiv preprint arXiv:2401.01339  (2024)

\bibitem{yang2023freenerf}
Yang, J., Pavone, M., Wang, Y.: Freenerf: Improving few-shot neural rendering with free frequency regularization. In: Proceedings of the IEEE/CVF Conference on Computer Vision and Pattern Recognition. pp. 8254--8263 (2023)

\bibitem{yang2023unisim}
Yang, Z., Chen, Y., Wang, J., Manivasagam, S., Ma, W.C., Yang, A.J., Urtasun, R.: Unisim: A neural closed-loop sensor simulator. In: Proceedings of the IEEE/CVF Conference on Computer Vision and Pattern Recognition. pp. 1389--1399 (2023)

\bibitem{yu2021pixelnerf}
Yu, A., Ye, V., Tancik, M., Kanazawa, A.: pixelnerf: Neural radiance fields from one or few images. In: Proceedings of the IEEE/CVF Conference on Computer Vision and Pattern Recognition. pp. 4578--4587 (2021)

\bibitem{yu2023mvimgnet}
Yu, X., Xu, M., Zhang, Y., Liu, H., Ye, C., Wu, Y., Yan, Z., Zhu, C., Xiong, Z., Liang, T., et~al.: Mvimgnet: A large-scale dataset of multi-view images. In: Proceedings of the IEEE/CVF Conference on Computer Vision and Pattern Recognition. pp. 9150--9161 (2023)

\bibitem{zhang2023control}
Zhang, L., Rao, A., Agrawala, M.: Adding conditional control to text-to-image diffusion models. In: Proceedings of the IEEE/CVF International Conference on Computer Vision. pp. 3836--3847 (2023)

\bibitem{zhang2018lpips}
Zhang, R., Isola, P., Efros, A.A., Shechtman, E., Wang, O.: The unreasonable effectiveness of deep features as a perceptual metric. In: Proceedings of the IEEE conference on computer vision and pattern recognition. pp. 586--595 (2018)

\bibitem{zhang2023completionformer}
Zhang, Y., Guo, X., Poggi, M., Zhu, Z., Huang, G., Mattoccia, S.: Completionformer: Depth completion with convolutions and vision transformers. In: Proceedings of the IEEE/CVF Conference on Computer Vision and Pattern Recognition. pp. 18527--18536 (2023)

\bibitem{zhou2023drivinggaussian}
Zhou, X., Lin, Z., Shan, X., Wang, Y., Sun, D., Yang, M.H.: Drivinggaussian: Composite gaussian splatting for surrounding dynamic autonomous driving scenes. arXiv preprint arXiv:2312.07920  (2023)

\bibitem{zhou2023sparsefusion}
Zhou, Z., Tulsiani, S.: Sparsefusion: Distilling view-conditioned diffusion for 3d reconstruction. In: Proceedings of the IEEE/CVF Conference on Computer Vision and Pattern Recognition. pp. 12588--12597 (2023)

\end{thebibliography}

\clearpage

% \clearpage
\section*{Appendix}
\appendix

% \section{Appendix}
% \label{sec:appendix}

In this appendix, we provide additional details omitted from the main manuscript due to the limited space. First, we present additional figures to underscore the motivation behind our proposed method (\cref{subsec:app-mo}). Then we present more implementation details on fine-tuning Diffusion Model\cite{rombach2022sd} and training 3DGS\cite{3dgs} (\cref{subsec:app-imple}). We also explore the influence of the Diffusion Model's prior on the generated results through dedicated experiment (\cref{subsec:app-exp}). Finally, we showcase more rendering results on the KITTI\cite{geiger2015kitti} and KITTI-360\cite{liao2022kitti} datasets (\cref{subsec:app-render}).

% \subsection{Preliminaries}
\vspace{-0.2cm}

\section{Motivation}
\label{subsec:app-mo}
\begin{figure}[htb]
  \centering
  \includegraphics[width=0.9\linewidth]{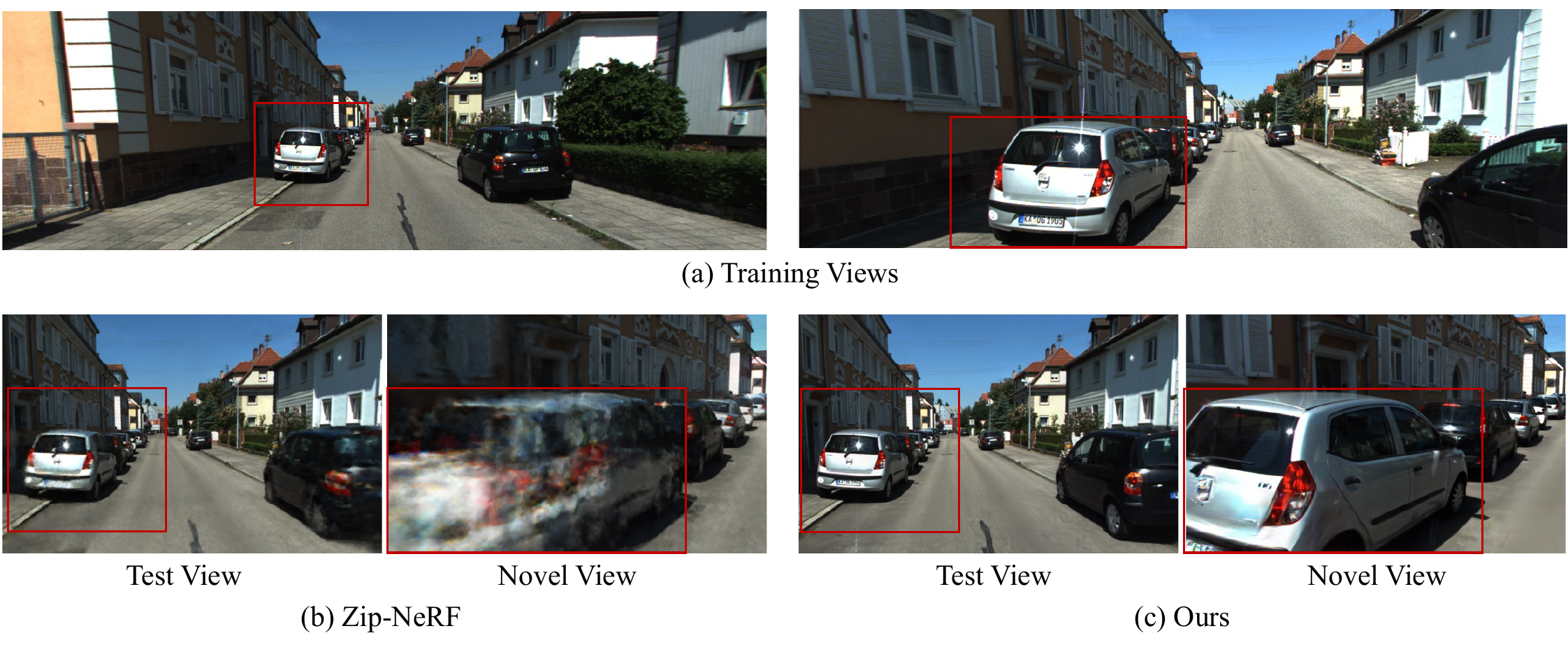}
  \caption{An example of how the current method\cite{barron2023zipnerf} overfits the training views, while our method overcomes this problem.}
  \label{fig:app-bg}
\end{figure}

Novel View Synthesis (NVS) for autonomous driving scenarios is a challenging task. The ideal training images for both NeRF\cite{mildenhall2021nerf} and 3DGS\cite{3dgs} should encompass all possible perspectives of the scene, which exhibit considerable disparities with the data collected by moving vehicles. The viewpoints offered by a vehicle-mounted camera are quite constrained. Take the white car in \cref{fig:app-bg} as an example, it is only observed from its side rear in the training view, causing the rendering model to overfit these viewpoints. While the current approach, such as Zip-NeRF\cite{barron2023zipnerf}, is able to render the vehicle clearly from a test view close to the training view, it produces unsatisfactory artifacts and deformation when the rendering viewpoint is shifted by a certain distance and rotated by a certain angle.

\vspace{-1cm}

\section{More Implementation Details}
\label{subsec:app-imple}
%CFG
% \subsubsection{Reference Images and Depth} 
\subsubsection{Diffusion Model} Our Diffusion Model is adapted from Stable Diffusion 1.5\cite{rombach2022sd} and is fine-tuned on about 12,000 images with $512\times512$ resolution from the KITTI-360\cite{liao2022kitti} dataset. Considering the original size of KITTI-360 images is $1408\times376$, a preliminary cropping step to $600\times376$ is performed before the resizing, to avoid over-distorting the images. We conduct \emph{center-crop} on the training images. For the reference images, we use \emph{random-crop} during the training process, which could ensure a certain perspective gap exists between the reference image and the training image, so as to enhance the robustness of the model. During inference, the reference images are pre-processed with \emph{center-crop}.

When selecting the reference images, we randomly choose one image from the five frames preceding the training image and one from the five frames succeeding the training image separately. During inference for the novel viewpoint, we identify its closest training viewpoint and utilize its adjacent frames as reference images.
Regarding the depth maps, due to the limitation of LiDAR point clouds in capturing the scene above a certain height, we apply a mask to the top 80 rows of pixels in the images. In practice, we found that the inpainting capability of the Stable Diffusion Model is effectively able to complete this portion of content. To enable classifier-free guidance in the first training stage, we set both text prompts and reference images to be empty with a 10\% probability.

\vspace{-0.2cm}

\subsubsection{3D Gaussian Splatting} We only initialize the 3D Gaussian models with LiDAR point cloud. The detailed procedure involves first projecting LiDAR frame onto its corresponding image frame to assign a color to each LiDAR point. Then these points are re-projected into 3D space, creating colored 3D point clouds. Finally, all frames of point clouds are accumulated and then voxel-downsampled with the voxel-size of 5. We train both our model and the baseline 3DGS model for 50,000 iterations. We first train the model for 500 iterations without sampling pseudo views for adequate warm-up. Subsequently, for every 10 iterations, 4 pseudo views are sampled for training.

\vspace{-0.2cm}

\section{Additional Experiment}
\label{subsec:app-exp}

As described in Sec. 3.2 of the main manuscript, during the training stage of 3DGS, we render some randomly sampled pseudo views, and utilize a fine-tuned Diffusion Model to generate guidance images for these views to regularize the training. Specifically, the pseudo view rendered by 3DGS is passed through the VAE Encoder to obtain a latent feature map, to which noise at level $t$ is added, where $t \sim [t_\text{min}, t_\text{max}]$. This noised latent feature is denoised by the Diffusion model from level $t$ to $t_\text{min}$, and then it is decoded to obtain the generated image. Specifically, we set $t_\text{max}=10$, and employ a hyper-parameter $s$, which indicates strength, to control the noise level $t$, according to $t = s \times t_\text{max}$. 

\begin{figure}[htb]
  \centering
  \includegraphics[width=0.98\linewidth]{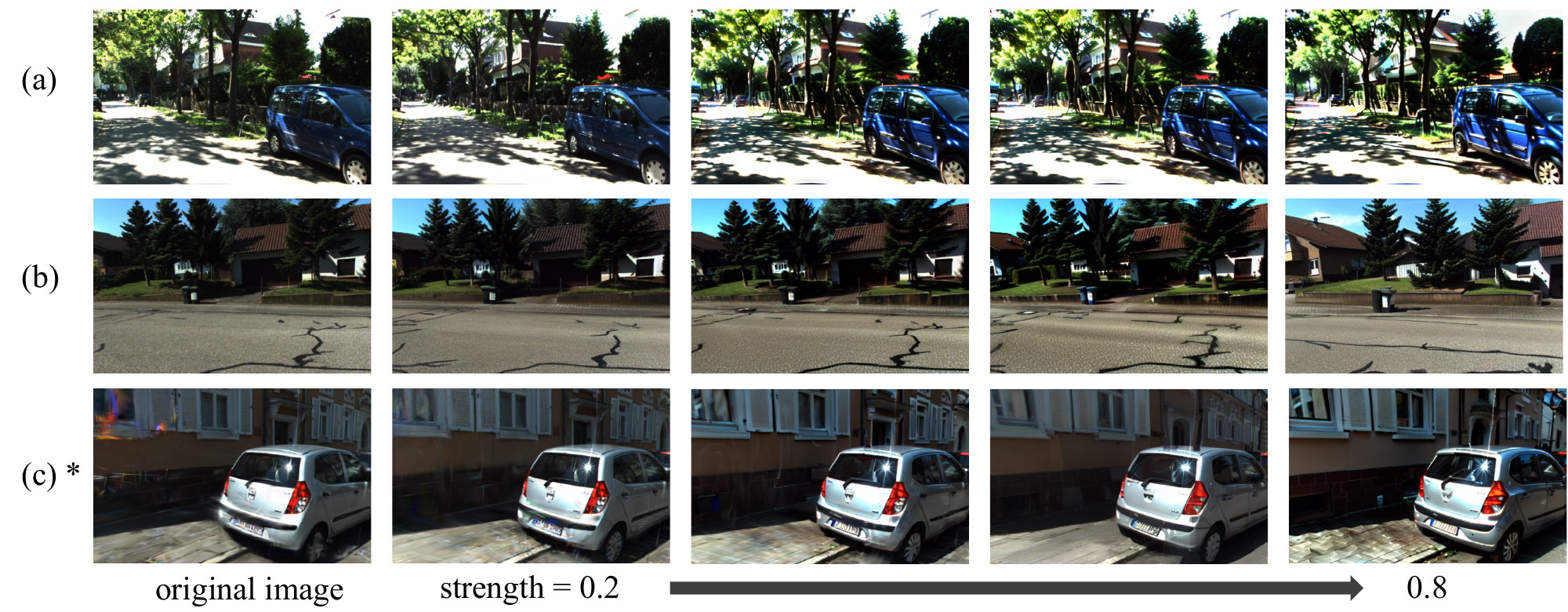}
  \caption{\textbf{The impact of the strength of the Diffusion Model's prior on the generated result.} *(c) is a novel view, its original image is rendered by 3DGS.}
  \label{fig:app-ablation}
\end{figure}

In \cref{fig:app-ablation}, we show the results of ablation experiments on hyper-parameter $s$. The first column labeled with \emph{original image} refers to the image being fed into the Diffusion Model, while the generated image with hyper-parameter $s$ increasing from 0.2 to 0.8 are exhibited in the other columns. It can be observed that a smaller $s$ makes generated images more similar to the original image, while a large $s$ introduces higher diversity and deviation in details. For novel viewpoints in \cref{fig:app-ablation}(c), smaller $s$ makes the generated image preserve noise rendered by 3DGS. As $s$ increases, the image becomes cleaner but loses some details. In practice, we randomly select $s \sim [s_\text{min}, s_\text{max}]$ for each sampled pseudo view, where $s_\text{min}=0.2$, $s_\text{max}$ starts at 0.6 and decreases to 0.4 over the training process. This strategy guarantees when 3DGS-rendered images are of lower quality in the early stage of training, our model relies more on the guidance from the Diffusion Model's prior. Accompanied by the quality of 3DGS renderings improves with ongoing training, it is necessary to reduce the impact of the Diffusion Model-generated images on the details.

% \vspace{-0.2cm}

\vspace{-0.2cm}

\section{More Rendering Results}
\label{subsec:app-render}

\begin{figure}[h]
  \centering
  \includegraphics[width=0.98\linewidth]{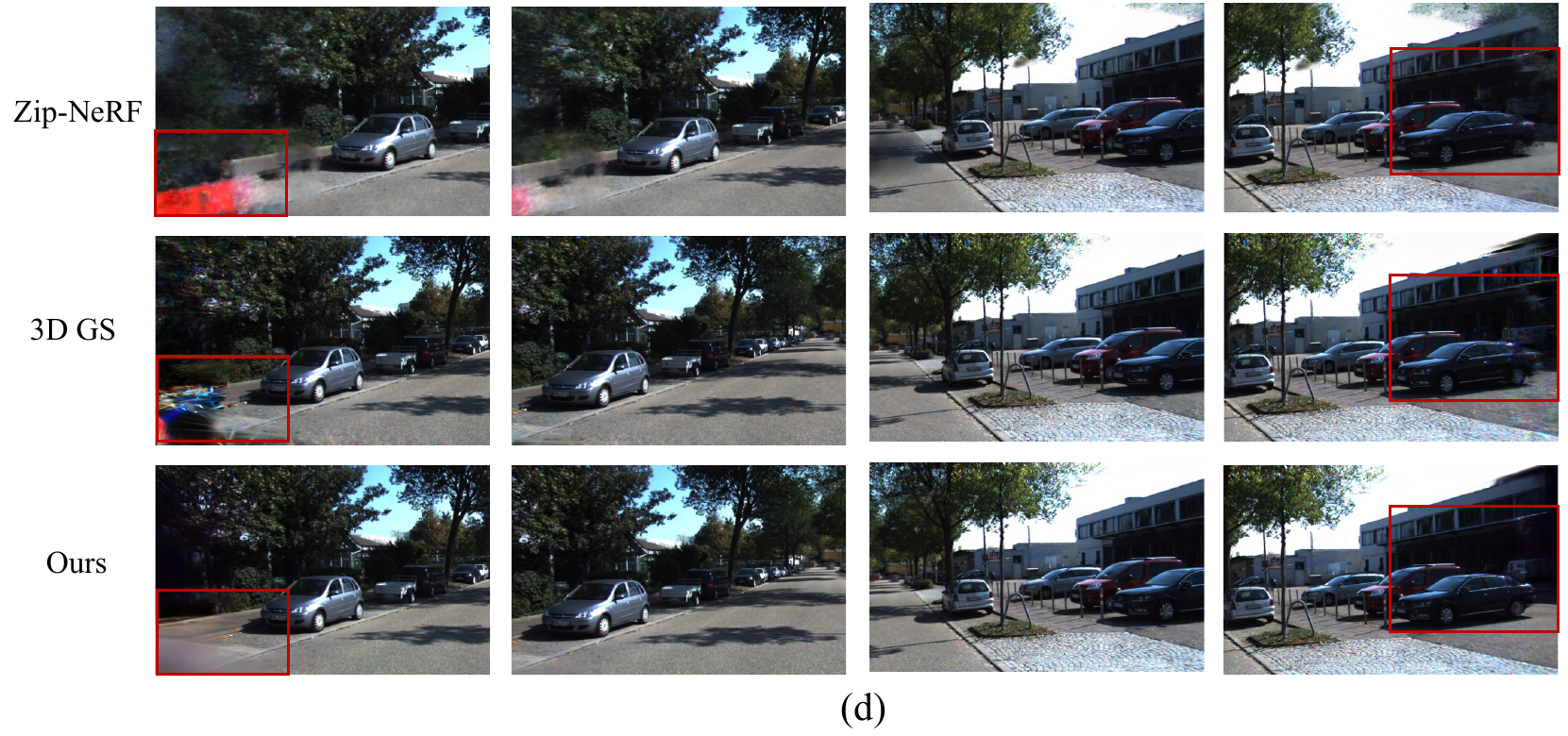}
  \caption{\textbf{More qualitative results of novel views rendering on the KITTI\cite{geiger2015kitti} and KITTI-360\cite{liao2022kitti} dataset.}}
  \label{fig:app-novel-e}
\end{figure}

We provide more novel view rendering results of our method and our competitors\cite{3dgs,barron2023zipnerf} on the KITTI\cite{geiger2015kitti} and KITTI-360\cite{liao2022kitti} datasets in \cref{fig:app-novel-e}.

\vspace{-0.2cm}

% \begin{figure}[!h]
%   \centering
%   \includegraphics[width=0.95\linewidth]{figs/app-kitti.pdf}
%   \caption{\textbf{More qualitative results of novel views rendering on KITTI\cite{geiger2015kitti}.}
%   }
%   \label{fig:app-kitti}
% \end{figure}

% \begin{figure}[htb]
%   \centering
%   \includegraphics[width=0.95\linewidth]{figs/app-360.pdf}
%   \caption{\textbf{More qualitative results of novel views rendering on KITTI-360\cite{liao2022kitti}.}
%   }
%   \label{fig:app-360}
% \end{figure}

\begin{figure}[h]
  \centering
  \includegraphics[width=0.98\linewidth]{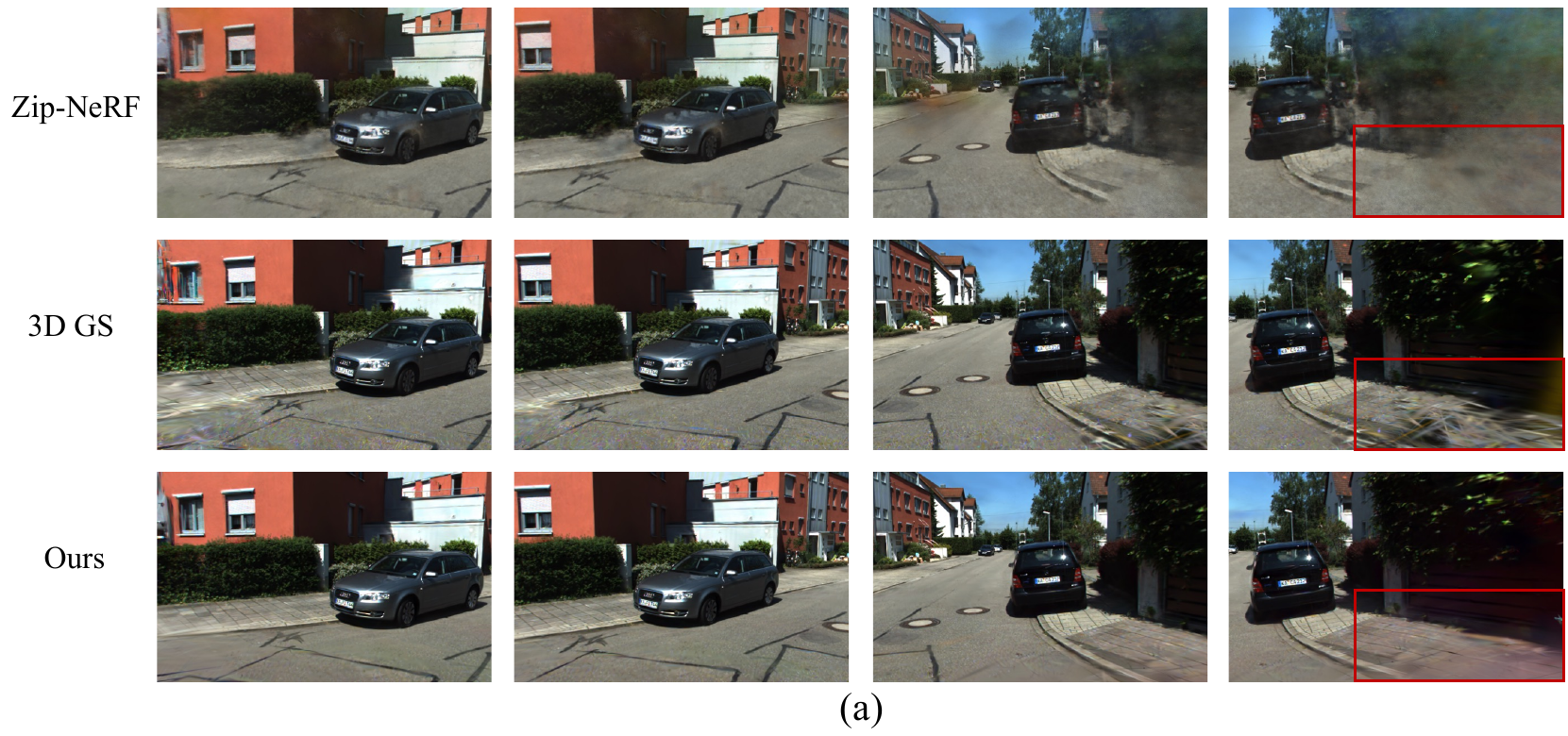}
  \label{fig:app-novel-a}
\end{figure}

\begin{figure}[h]
  \centering
  \includegraphics[width=0.98\linewidth]{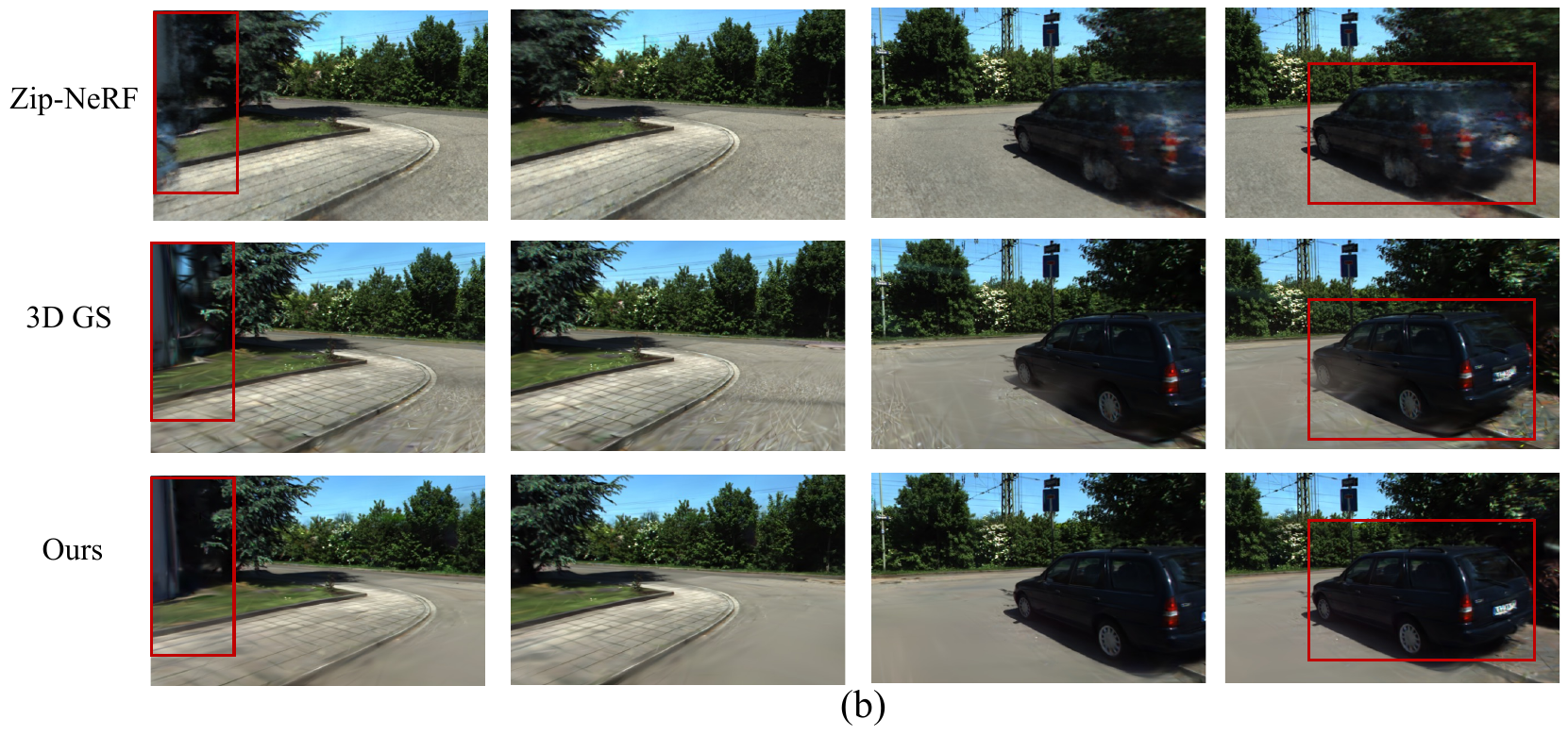}
  \label{fig:app-novel-b}
\end{figure}

% \begin{figure}[h]
%   \centering
%   \includegraphics[width=0.98\linewidth]{figs/appendix_novels_c.pdf}
%   \label{fig:app-novel-c}
% \end{figure}

\begin{figure}[h]
  \centering
  \includegraphics[width=0.98\linewidth]{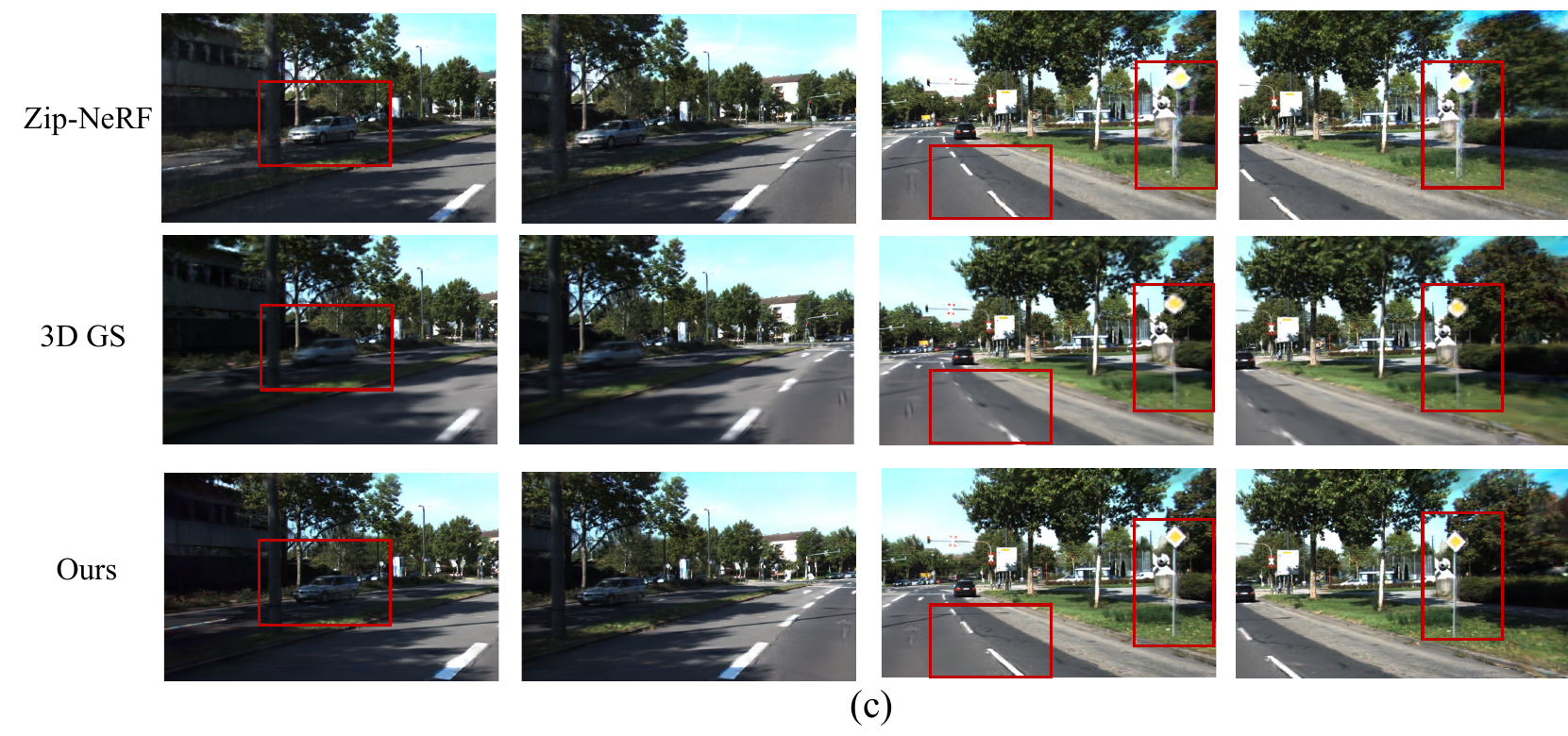}
  \label{fig:app-novel-d}
\end{figure}

\end{document}